\documentclass[fleqn,10pt]{wlscirep}
\usepackage[utf8]{inputenc}
\usepackage[T1]{fontenc}

\usepackage{colortbl}
\usepackage{xcolor} 
\usepackage{todonotes}
\usepackage{cleveref}
\usepackage{float}
\usepackage{multirow}

\usepackage{soul}
\usepackage{xcolor}
\soulregister{\cite}{1} 
\soulregister{\ref}{1}  

\usepackage{color}
\usepackage{soul} 
\usepackage{xcolor} 
\usepackage{amssymb}  
\usepackage{pifont}   
\usepackage{soul} 

\usepackage{soul} 
\usepackage{xcolor} 

\title{Domain-incremental White Blood Cell Classification with Privacy-aware Continual Learning}

\author[1,+,*]{Pratibha Kumari}
\author[1,+]{Afshin Bozorgpour}
\author[1]{Daniel Reisenbüchler}
\author[4]{Edgar Jost}
\author[4]{Martina Crysandt}
\author[3]{Christian Matek}
\author[1,2]{Dorit Merhof}

\affil[1]{University of Regensburg, 93053, Germany}
\affil[2]{Fraunhofer Institute for Digital Medicine MEVIS, Bremen, Germany}
\affil[3]{University Hospital Erlangen, Erlangen, Germany}
\affil[4]{Department of Hematology, Oncology, Hemostaseology and Stem Cell Transplantation, University Hospital RWTH Aachen, Aachen, Germany}

\affil[*]{\centering corresponding author: \texttt{pratibha.kumari@ur.de}}

\affil[+]{\centering \small these authors contributed equally to this work}

\begin{abstract}

White blood cell (WBC) classification plays a vital role in hematology for diagnosing various medical conditions. However, it faces significant challenges due to domain shifts caused by variations in sample sources (e.g., blood or bone marrow) and differing imaging conditions across hospitals. Traditional deep learning models often suffer from catastrophic forgetting in such dynamic environments, while foundation models, though generally robust, experience performance degradation when the distribution of inference data differs from that of the training data. To address these challenges, we propose a generative replay-based Continual Learning (CL) strategy designed to prevent forgetting in foundation models for WBC classification. Our method employs lightweight generators to mimic past data with a synthetic latent representation to enable privacy-preserving replay. To showcase the effectiveness, we carry out extensive experiments with a total of four datasets with different task ordering and four backbone models including ResNet50, RetCCL, CTransPath, and UNI. Experimental results demonstrate that conventional fine-tuning methods degrade performance on previously learned tasks and struggle with domain shifts. In contrast, our continual learning strategy effectively mitigates catastrophic forgetting, preserving model performance across varying domains. This work presents a practical solution for maintaining reliable WBC classification in real-world clinical settings, where data distributions frequently evolve.
\end{abstract}
\begin{document}

\flushbottom
\maketitle
%
%
\thispagestyle{empty}

\section{Introduction}\label{sec:intro}

White blood cell (WBC) classification plays a pivotal role in hematology for diagnosing conditions such as infections, immune disorders, and leukemia. Advances in deep learning have revolutionized this field by extracting subtle patterns in complex samples that often elude human experts. The performance of these models has been significantly enhanced by leveraging large and varied training datasets. Moreover, self-supervised learning techniques have enabled the use of vast unannotated datasets, leading to the development of foundation models (FMs) tailored for medical data.

Traditional deep learning models for WBC classification typically assume a static training environment, expecting consistent data distributions between the training and testing phases. Although extensive datasets amassed over decades improve generalization, these models may falter in dynamic clinical settings where data distributions shift due to variations in sample sources, collection procedures, and imaging equipment across different hospitals. 
To deal with varying data distribution, a straightforward approach involves retraining the models from scratch whenever new data becomes available. While this can theoretically ensure alignment with the latest data distribution, it is both computationally expensive and resource-intensive, making it impractical for dynamic clinical environments with continuously evolving datasets. A simple alternative would be to fine-tune the existing model on new data. However, such a naive adaptation strategy tends to overwrite previously acquired knowledge, leading to catastrophic forgetting, a phenomenon where the model's performance on earlier tasks deteriorates significantly~{\cite{goodfellow2013empirical}}. Thus, proposing novel methods that can effectively incorporate new information without sacrificing existing knowledge is crucial for reliable, long-term deployment of WBC classification systems.

Continual Learning (CL) addresses this challenge by enabling models to assimilate new information while retaining prior knowledge~\cite{kumari2023continual,KUMARI2024117100,bhatt2022experimental}. 
Recent studies, such as by Sadafi et al.~{\cite{sadafi2023continual}}, have explored replay-based CL approaches for class-incremental WBC classification by storing a subset of past images to mitigate forgetting. However, storing raw medical images raises significant privacy concerns, limiting the practical application of such methods.

Motivated by the above-mentioned limitations, we propose a privacy-aware generative replay-based CL framework for WBC classification from both blood and bone marrow smears. Instead of storing images, our approach focuses on encapsulating past data distributions while preserving privacy.
To this end, we employ a non-parametric generator using Kernel Density Estimation (KDE) on latent vectors. We compared it with recently published similar work, Generative Latent Replay-based CL (GLRCL)~\cite{Kum_Continual_MICCAI2024}, which uses Gaussian Mixture Model (GMM) based generative replay. In contrast to the GLRCL approach, which assumes that the data follows a Gaussian distribution, our KDE-based generator eliminates the need to predetermine the number of components due to its non-parametric nature. It does not impose any assumptions about the underlying data distribution and provides greater flexibility in modeling arbitrary data distributions~\cite{gidel2010comparison}.

Our framework employs the KDE-based generator to create latent vectors that mimic past domains during training on new data. Furthermore, we incorporate a Kullback-Leibler divergence-based distillation mechanism to guide weight updates, ensuring that critical information from previous tasks is retained.
We evaluate our approach on four datasets exhibiting domain shifts arising from different collection centers and organ sources. Extensive experiments across multiple dataset orderings demonstrate that our approach outperforms several CL benchmarks. Furthermore, we compare various backbone architectures, including foundation models, highlighting the effectiveness of our approach in mitigating catastrophic forgetting and maintaining robust performance across domains.

Our main contributions can be summarized as follows:
\begin{itemize}
\item \textbf{WBC Classification in Domain Incremental Scenario}: We present the first study on WBC classification in domain-incremental scenarios, analyzing the impact of domain shifts in real-world datasets. Our work highlights the necessity of CL for updating FMs without requiring access to previously learned data.

\item \textbf{Privacy-Aware CL Approach}: We introduce a privacy-aware generative replay strategy for WBC classification using a KDE-based generator, eliminating the need to store raw medical images. Unlike existing CL methods that assume Gaussian-distributed data, our non-parametric approach provides greater flexibility in modeling diverse clinical distributions. Additionally, we integrate a KL-divergence-based distillation mechanism to prevent catastrophic forgetting while adapting to new domains.

\item \textbf{Comprehensive Evaluation:} We are among the first to explore CL strategies across diverse backbone architectures, including FMs. Our results demonstrate that the performance of FMs degrades under domain shifts, but integrating them with CL mitigates catastrophic forgetting, ensuring robust performance. We validate our approach on four WBC datasets (PBC, LMU, MLL, UKA) and multiple backbones (ResNet50, RetCCL, CTransPath, UNI), demonstrating its effectiveness in handling heterogeneous domain shifts.

\end{itemize}

\begin{figure*}[!ht]
    \centering
    \includegraphics[scale=0.99]{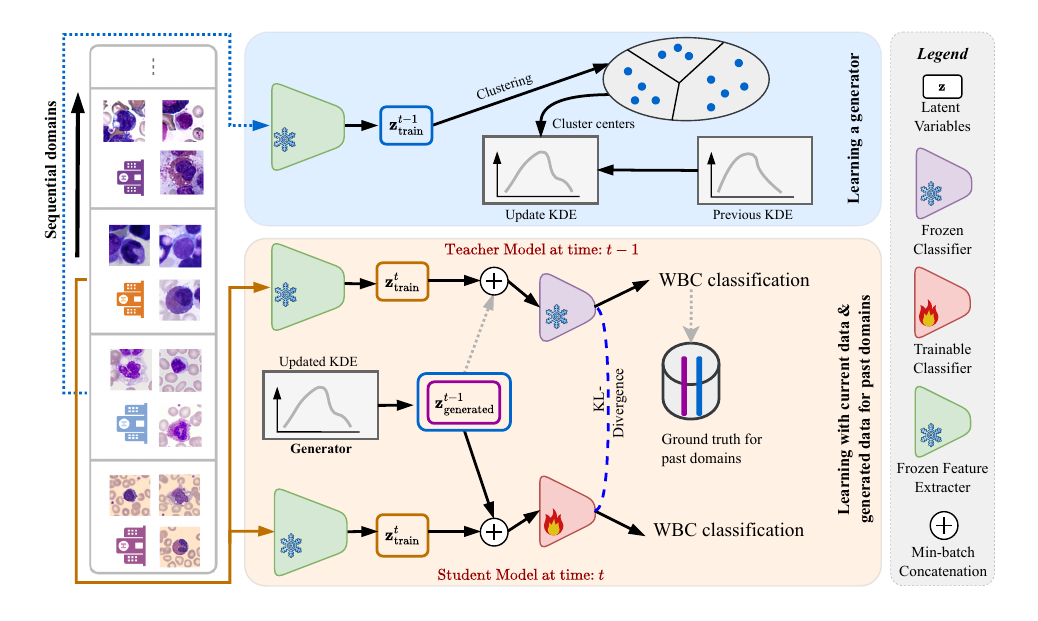}
    \caption{Flowchart of the proposed continual learning-based framework for WBC classification in heterogeneous domain shift conditions.}
    \label{fig:framework_WorkShop}
\end{figure*}



\section{Methodology}\label{sec:proposed}

We address the challenge of WBC classification in scenarios where datasets originate from various sources, including different hospitals and different types of samples (e.g., blood and bone marrow), leading to heterogeneous domain shifts. In CL, this setting is referred to as the domain-incremental scenario, where the target classes remain consistent across datasets, but the data distributions differ. In a domain incremental scenario of CL, we are given $T$ datasets, potentially from different sources, which need to be learnt in a sequence. The datasets in a sequence are also referred to as tasks or episodes. The training and testing pairs for the $T$ tasks are given by \(\{D^{1}_{\text{train}}, D^{1}_{\text{test}}\}, \{D^{2}_{\text{train}}, D^{2}_{\text{test}}\}, \dots, \{D^{T}_{\text{train}}, D^{T}_{\text{test}}\}\), where \(\{D^{t}_{\text{train}}, D^{t}_{\text{test}}\}\) represents the training and testing data for the \(t^{\text{th}}\) task.
During training on the \(t^{\text{th}}\) task, access to all prior tasks' data ($<t$) is limited or nonexistent, requiring the model to learn sequentially in total $T$ training sessions. Our CL model aims to optimize performance across all tasks at each training session.

A visual summary of our proposed CL based WBC classification framework is shown in Fig. {\ref{fig:framework_WorkShop}}. The proposed framework features two core components: (a) a non-parametric latent generator to mimic past tasks (Section 2.1) and (b) a generative latent replay and knowledge distillation strategy to counteract catastrophic forgetting (Section 2.2).

\subsection{Non-parametric Latent Generator}
The latent generator is designed to be both lightweight and privacy-aware, eliminating the need to store sensitive raw WBC images. For the current training session $t$, discriminative latent features $\mathbf{z}^{t}_{\text{train}}$ are extracted from $D^{t}_{\text{train}}$ using a pretrained deep neural network backbone. To compactly represent these features, K-means clustering is applied with $K$ clusters. The Bayesian Information Criterion (BIC) is used to determine the optimal number of clusters, ensuring a balance between complexity and representational accuracy. Only the resulting cluster centers are retained, capturing the essential data distribution while discarding non-essential features to enhance computational efficiency. 
These centers are used to update an existing KDE-based generator $G^{t-1}$ which represents the generator for all older tasks ${\{1,\dots,t-1\}}$. Basically, we combine latent points from the previous generator $G^{t-1}$ and cluster centers from the $t^{\text{th}}$ task to obtain the updated generator $G^{t}$.
The bandwidth parameter $B$ of our KDE-based generator is automatically determined using Silverman's rule of thumb, allowing the generator to adapt to the data's inherent variability. By progressively updating the generator after each task, the generator can be used to effectively simulate all past data distributions without retaining original samples.

\subsection{Forgetting Control}
To prevent catastrophic forgetting, our framework employs two complementary strategies: generative latent replay and knowledge distillation-based regularization.

\noindent
\textbf{Generative Latent Replay}: 
This approach enables the model to retain prior knowledge by synthesizing representative latent vectors from past tasks instead of storing raw WBC data. During the training of the $t^{th}$ task, a classification model $M^t$ (considered as student model) is initialized from the immediate trained past model $M^{t-1}$ (considered as teacher model). The training process of $M^t$ employs a hybrid mini-batch that includes samples from the current task data $D^{t}_{\text{train}}$ as well as generated latent vectors $\mathbf{z}^{t-1}_{\text{generated}}$, which are produced by the KDE-based generator $G^{t-1}$ to represent all past tasks. Each mini-batch is composed of an equal split: 50\% of the samples originate from the current task, while the remaining 50\% are synthetic samples representing data from previous tasks. Since the generated latent vectors do not have explicit labels, the teacher model $M^{t-1}$, which has already learned these past tasks, is used to assign pseudo-labels through its predictions, as illustrated by the dotted lines in Fig.~{\ref{fig:framework_WorkShop}}. This process ensures that both previous as well as current information is effectively integrated at the feature level, thus preventing catastrophic forgetting while maintaining the privacy of past data.

\noindent
\textbf{Knowledge Distillation}:
To further enhance learning stability, our framework incorporates a knowledge distillation term into the training loss function~{\cite{kumari2023continual}}. This regularization strategy encourages the student model $M^{t}$ to retain essential information from the previously trained teacher model $M^{t-1}$. Specifically, we employ the Kullback-Leibler (KL) divergence to align the output logits of the student and teacher models. The overall training loss is defined as:
\begin{equation}
\text{loss} = (1 - \alpha) \times \text{loss}_{CE} + \alpha \times \text{loss}_{KLD},
\end{equation}
where $\text{loss}_{CE}$ represents the cross-entropy loss between the student model's predictions and the ground truth labels, and $\text{loss}_{KLD}$ denotes the KL divergence between the logits of $M^t$ and $M^{t-1}$. The parameter $\alpha \in [0,1]$ serves as a balancing factor, controlling the trade-off between learning new information and retaining previously acquired knowledge. This approach ensures that the student model maintains consistency with earlier learned representations, thereby mitigating catastrophic forgetting as new tasks are introduced.

\begin{table*}[!ht]
    \centering
    \caption{Details of datasets}
    \label{tab:datasetTable}
    \def\arraystretch{0.9}
    \setlength{\tabcolsep}{5pt}
    \begin{tabular}{|c|c|c|c|c|}
    \hline
\rowcolor{gray!10}    \begin{tabular}[c]{@{}c@{}} Dataset  \end{tabular}& Center & Source  & \begin{tabular}[c]{@{}c@{}}\#Train samples \\Classes: MON, EOS, LYT \end{tabular} & \begin{tabular}[c]{@{}c@{}}\#Test samples \\ Classes: MON, EOS, LYT \end{tabular} \\ \hline

        PBC~\cite{acevedo2020dataset}& $C1$ & Blood smears & 1370, 3067, 1164 & 50, 50, 50 \\ 
      
       LMU~\cite{matek2019single} & $C2$ & Blood smears & 1739,  374, 3887 & 50, 50, 50 \\
        
        MLL~\cite{matek2021expert} & $C3$ & Bone-marrow  & 3990, 5833, 6000 & 50, 50, 50 \\ 
        
        UKA & $C4$ & Bone-marrow  &  239,  124,  229 & 50, 50, 50 \\ 
    \hline
    \end{tabular}
\end{table*}

\begin{table*}[!ht]
    \centering
    \caption{Details of training dataset sequences used in experiments}
    \label{tab:seqDatasetTable}
    \def\arraystretch{0.9}
    \setlength{\tabcolsep}{5pt}
    \begin{tabular}{|c|c|l|}
    \hline
 \rowcolor{gray!10}   \begin{tabular}[c]{@{}c@{}} Name of sequence \end{tabular}& Dataset order & Order based on \\ \hline
        Seq.1 & PBC, LMU, MLL, UKA & blood smears followed by bone-marrow datasets \\ 
        Seq.2 & MLL, PBC, UKA, LMU & bone-marrow and blood smears are interleaved \\ 
        Seq.3 & UKA, MLL, LMU, PBC & bone-marrow followed by blood smears \\ 
        Seq.4 & UKA, PBC, LMU, MLL & low to high training data \\ 
    \hline
    \end{tabular}
\end{table*}

\section{Experimental Setup}\label{sec:exp}
\subsection{Datasets}\label{sec:datasets}

We utilize a series of WBC datasets demonstrating challenging domain shifts commonly found in clinical settings, which we refer to as `heterogeneous shifts'. These shifts can be attributed to differences in acquisition centers, staining protocols, imaging conditions, and organs (blood smears and bone marrow). We consider four datasets, namely PBC, LMU, MLL, and UKA, which exhibit notable variations.
\\ \noindent
\textbf{PBC}: This dataset consists of 17,092 images of White Blood Cells (WBCs), which are categorized into 11 distinct classes. This dataset is derived from normal peripheral blood cells~\cite{acevedo2020dataset}. The smears were prepared with May Grünwald-Giemsa staining using the slide maker-stainer Sysmex SP1000i. The cell images were digitized with CellaVision DM96 to JPG images with size 360 × 363 pixels.
\\ \noindent
\textbf{LMU}: As presented in~\cite{matek2019single}, this dataset contains over 18,000 images of single WBCs from peripheral blood smears that have been classified into 5 major types~\cite{matek2019single}. The single-cell images were produced by M8 digital microscope/scanner (Precipoint GmbH, Freising/Germany) at 100$\times$ magnification and oil immersion. Here, the physical size of a camera pixel is 14.14 pixels per $\mu$m.
\\ \noindent
\textbf{MLL}: This dataset is a subset from another publicly available source~\cite{matek2021expert}. It consists of bone marrow smears from cohorts over 170,000 images of WBCs, encompassing 8 different cell types stained at the Munich Leukemia Laboratory. The bone marrow smears were scanned with a magnification of 40$\times$, stained with May-Grünwald-Giemsa/Pappenheim staining. The samples were digitized with a Zeiss Axio Imager Z2 into 2452 × 2056 pixels, and the physical size of a camera pixel is 3.45 × 3.45 $\mu$m.
\\ \noindent
\textbf{UKA}: This is an internal WBC dataset derived from bone marrow smears, comprising 11,899 labeled images of WBCs across 36 distinct classes, collected from 6 patients at Uniklinikum Aachen. The samples were acquired using a Leica Biosystems Aperio VERSA 200 whole-slide scanner with 63$\times$ magnification and oil immersion. The physical size of a camera pixel is approximately 11.5 pixels per $\mu$m.

Given that these datasets originate from different clinical sites, we refer to the respective clinical centers as $C1$, $C2$, $C3$, and $C4$. As mentioned in Section~{\ref{sec:proposed}}, for a domain incremental CL scenario, a fixed number of classes is required across all the episodes. We consider monocyte (MON), eosinophil (EOS), and lymphocyte (LYT) classes in each dataset. Classes other than MON, EOS, and LYT are excluded due to their limited availability in some datasets. More details regarding class selection are provided in the Supplements. Details regarding the number of samples, centers, and organs for each dataset are provided in \Cref{tab:datasetTable}. For an extensive evaluation, we create multiple CL experiments with the above-mentioned datasets. Specifically, we form four different sequences on these datasets, varying the dataset order in terms of increasing or decreasing dataset volume, as well as organ-specific learning. We mention the name of each of these four sequences, dataset order in each sequence, and the basis of ordering in \Cref{tab:seqDatasetTable}.

\subsection{Baselines}
We compare our method against various CL benchmarks, including regularization-based approaches which do not keep memory buffers, but add a regularization term in the training loss, such as EWC~\cite{kirkpatrick2017overcoming}, SI~\cite{zenke2017continual}, and LwF~\cite{radio3}. Furthermore, we assess replay-based approaches that utilize buffers, including DER~\cite{buzzega2020dark}, ER~\cite{rolnick2019experience}, LR~\cite{pellegrini2020latent}, GEM~\cite{lopez2017gradient}, and AGEM~\cite{chaudhry2018efficient}. We also compare with GLRCL, a recent CL framework employing GMM-based generative replay for tumor classification in domain shift conditions \cite{Kum_Continual_MICCAI2024}. 
Along with CL approaches, we also report the performances of non-CL methods, namely, the naive and joint strategy. In the naive approach, the model is fine-tuned using only the new task data without any mechanism to mitigate forgetting, thereby establishing a lower bound for the average performance across all tasks. In contrast, joint training assumes that all tasks are simultaneously available, enabling the model to train on all tasks concurrently and achieve an upper performance bound.

\subsection{Classification Models} 
Primarily, ImageNet-pretrained models like ResNet50 and its variants have been utilized for CL-based classification frameworks. However, the significant domain gap between natural and medical data may limit CL performance. To address this, we also validate for SSL-based models specifically tailored for pathology data. To this end, we used RetCCL~\cite{wang2023retccl}, a ResNet50 trained on a large cohort of human cancer pathology images. CTransPath~\cite{wang2022transformer} and UNI~\cite{chen2024towards} are transformer based FMs with more capacity and trained on multi-scale with larger amount of data. DinoBloom~\cite{koch2024dinobloom}, an FM for hematology is not considered in our work as this backbone has been trained on variety of WBC datasets including the ones used in our experiments (Section~\ref{sec:datasets}). To construct our classification model, we select a backbone from \{ResNet50, RetCCL, CTransPath, UNI\}, discard the respective classification head and appended five learnable Fully Connected (FC) layers $[512, 256, 128, 64, 32]$ for supervised classification. Each backbone is initialized with pre-trained weights from its respective repository. Following the approach in {\cite{Kum_Continual_MICCAI2024}}, we freeze the backbone weights to preserve the KDE-based latent generator's validity, training only the FC layers.

\subsection{Training Setup} 
We followed standard CL evaluation protocols using the Avalanche library~\cite{avalanche}. The WBC images were resized to \(256 \times 256\) pixels, and batched with a size of 64. 
For our approach, we set best \(\alpha\) (regularization coefficient) by following grid search in set \{0.01, 0.1, 0.2, 0.3, 0.4, 0.5\} for each backbone. Similar to work~\cite{Kum_Continual_MICCAI2024}, we set the number of cluster centers for a domain to $10$ in K-means clustering. Thus, the number of total cluster centers on which KDE is built is $40$ latent vectors ($10\times 4$ for 4 tasks in a sequence). 
For fair comparison with buffer based approaches we keep buffer size in GEM, AGEM, ER, LR, and DER as $40$. 
We followed~\cite{Kum_Continual_MICCAI2024} and set the number of components for per-domain GMM to $10$. For SI, EWC, and LwF, the value of regularizing factor ($\alpha$) was set to 1 by following the literature~\cite{Kum_Continual_MICCAI2024}. All experiments were conducted using an NVIDIA A100 GPU, with training and evaluation sessions completing within a maximum of two hours across all methods, except for GEM which took a maximum of 15 hours. To ensure statistical reliability, all experiments were repeated five times, and we report the mean and standard deviation of the results.

\subsection{Evaluation Metrics} 
After sequentially learning \(T\) classification tasks, we constructed a train-test matrix \cite{kumari2023continual} \(P\), where each cell \(p_{ij}\) represents the accuracy on $j^{\text{th}}$ dataset after completing training on $i^{\text{th}}$ dataset. From this matrix, we computed CL metrics such as backward transfer (BWT) \cite{diaz2018don}, average accuracy after completing the last task (ACC) \cite{lopez2017gradient}, and Incremental Learning Metric (ILM) \cite{diaz2018don}, which measures the incremental learning capability of the approach. Higher values of these metrics indicate better performance. We detail the equations for these metrics in Table~{\ref{tab:equationTable}}.

\begin{table*}[!htbp]
\centering
\renewcommand{\arraystretch}{1.5}
\begin{tabular}{|c|c|c|c|}
\bottomrule
\cellcolor{gray!10}
{\bf Metric Name} &  {BWT} & {ACC} & {ILM} \\
\hline
\cellcolor{gray!10}
{\bf Equation} &
$\frac{1}{T-1}\sum_{j=1}^{T-1}\frac{1}{\left |\left \{  t_i\right \}_{i>j} \right |}\sum_{i>j}^{}\left [ p_{i,j}- p_{j,j} \right ]$ &
$\frac{1}{T}\sum_{j=1}^{T} p_{T,j}$ &
$\frac{2}{T (T+1)}\sum_{j \leq i} p_{i,j}$ \\
\toprule
\end{tabular}
\caption{Equations for BWT, ACC, and ILM metrics.}
\label{tab:equationTable}
\end{table*}
\begin{figure}[!ht]
\centering
\includegraphics[width=\textwidth]{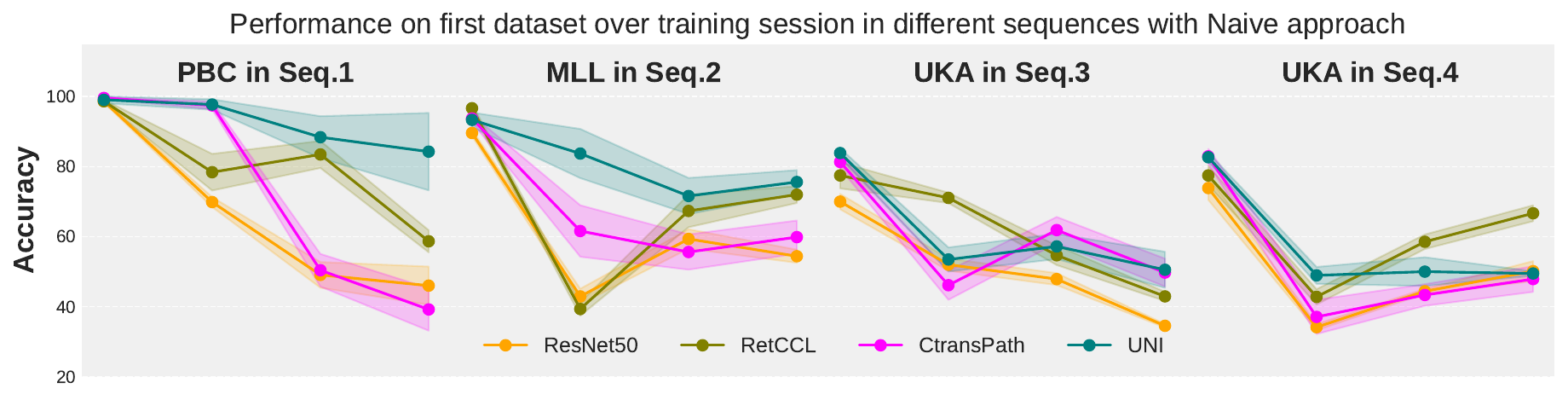} \\
\includegraphics[width=\textwidth]{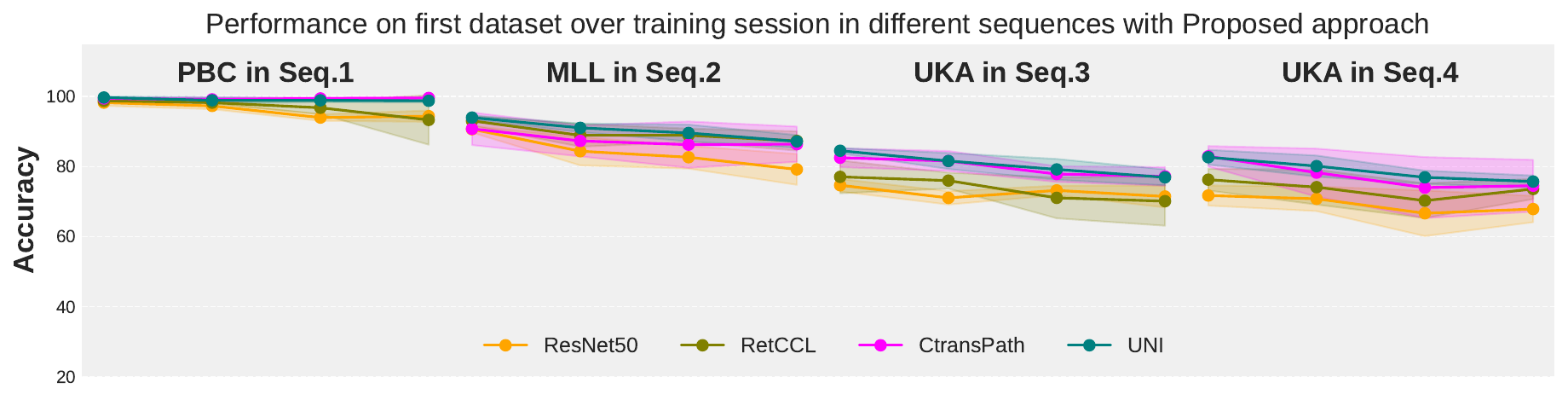} \\
\caption{Performance of the first dataset upon learning other subsequent datasets in Seq.1 to Seq.4 in a naive (top row) and proposed (bottom row) way}
\label{fig:naiveANDkdeResultsOrganWise}
\end{figure}
\begin{table*}[t]
\centering

\caption{ Performance comparison under \textbf{ResNet50} backbone (\textcolor{red}{best result}, \textcolor{blue}{second best result} in continual learning methods)}
 \label{tab:comp_cl_all_ResNet}
\resizebox{\textwidth}{!}{%
\def\arraystretch{0.9}
\setlength{\tabcolsep}{5pt}
\begin{tabular}{|l|ccc|ccc|ccc|ccc|ccc|}
\hline
\rowcolor{gray!10} 
 & \multicolumn{3}{c|}{\textbf{Seq.1}} & \multicolumn{3}{c|}{\textbf{Seq.2}} & \multicolumn{3}{c|}{\textbf{Seq.3}} & \multicolumn{3}{c|}{\textbf{Seq.4}} & \multicolumn{3}{c|}{\textbf{Average}} \\ \hline
\rowcolor{gray!10} 
\begin{tabular}[c]{@{}c@{}}\textbf{Method} \end{tabular} & \textbf{ACC}$\uparrow$ & \textbf{ILM}$\uparrow$ & \textbf{BWT}$\uparrow$ &
\textbf{ACC}$\uparrow$ & \textbf{ILM}$\uparrow$ & \textbf{BWT}$\uparrow$ &
\textbf{ACC}$\uparrow$ & \textbf{ILM}$\uparrow$ & \textbf{BWT}$\uparrow$ & 
\textbf{ACC}$\uparrow$ & \textbf{ILM}$\uparrow$ & \textbf{BWT}$\uparrow$ &
\textbf{ACC}$\uparrow$ & \textbf{ILM}$\uparrow$ & \textbf{BWT}$\uparrow$ \\
\hline

Naive  &56.07 {\hspace{-5pt} \tiny $\pm 1.53$} & 67.83 {\hspace{-5pt} \tiny $\pm 1.44$} & -39.13 {\hspace{-5pt} \tiny $\pm 2.54$}  &66.07 {\hspace{-5pt} \tiny $\pm 1.22$} & 67.53 {\hspace{-5pt} \tiny $\pm 0.34$} & -36.24 {\hspace{-5pt} \tiny $\pm 0.97$}  &59.77 {\hspace{-5pt} \tiny $\pm 0.68$} & 64.01 {\hspace{-5pt} \tiny $\pm 0.47$} & -29.33 {\hspace{-5pt} \tiny $\pm 1.71$}  &58.83 {\hspace{-5pt} \tiny $\pm 1.57$} & 64.45 {\hspace{-5pt} \tiny $\pm 0.82$} & -35.71 {\hspace{-5pt} \tiny $\pm 2.19$}
&  60.18 {\hspace{-5pt} \tiny $\pm 0.36$}  & 65.96 {\hspace{-5pt} \tiny $\pm 0.43$}  & -35.1 {\hspace{-5pt} \tiny $\pm 0.59$} \\ 

Joint
& 88.60 {\hspace{-5pt} \tiny $\pm 0.42$}
&-&-& 87.73 {\hspace{-5pt} \tiny $\pm 0.96$}
&-&-& 87.83 {\hspace{-5pt} \tiny $\pm 0.89$}
&-&-&87.47 {\hspace{-5pt} \tiny $\pm 0.61$}
&-&-&87.91{\hspace{-5pt} \tiny $\pm 0.49$}&-&-\\\hline

DER  &56.37 {\hspace{-5pt} \tiny $\pm 3.25$} & 67.68 {\hspace{-5pt} \tiny $\pm 1.08$} & -38.07 {\hspace{-5pt} \tiny $\pm 2.89$}  &66.67 {\hspace{-5pt} \tiny $\pm 1.21$} & 67.51 {\hspace{-5pt} \tiny $\pm 0.77$} & -37.0 {\hspace{-5pt} \tiny $\pm 1.95$}  &57.93 {\hspace{-5pt} \tiny $\pm 1.17$} & 63.17 {\hspace{-5pt} \tiny $\pm 0.99$} & -32.55 {\hspace{-5pt} \tiny $\pm 2.2$}  &61.7 {\hspace{-5pt} \tiny $\pm 2.42$} & 66.12 {\hspace{-5pt} \tiny $\pm 1.0$} & -31.82 {\hspace{-5pt} \tiny $\pm 2.72$}
&  60.67 {\hspace{-5pt} \tiny $\pm 0.87$}  & 66.12 {\hspace{-5pt} \tiny $\pm 0.12$}  & -34.86 {\hspace{-5pt} \tiny $\pm 0.38$} \\

 ER &66.57 {\hspace{-5pt} \tiny $\pm 3.35$} & 76.41 {\hspace{-5pt} \tiny $\pm 1.33$} & -24.74 {\hspace{-5pt} \tiny $\pm 2.04$}  &71.4 {\hspace{-5pt} \tiny $\pm 1.32$} & 76.76 {\hspace{-5pt} \tiny $\pm 0.59$} & -20.13 {\hspace{-5pt} \tiny $\pm 1.24$}  &66.27 {\hspace{-5pt} \tiny $\pm 3.28$} & 70.16 {\hspace{-5pt} \tiny $\pm 1.82$} & -20.93 {\hspace{-5pt} \tiny $\pm 2.41$}  &70.84 {\hspace{-5pt} \tiny $\pm 2.74$} & 73.73 {\hspace{-5pt} \tiny $\pm 1.76$} & -19.51 {\hspace{-5pt} \tiny $\pm 3.4$}
&  68.77 {\hspace{-5pt} \tiny $\pm 0.82$}  & 74.26 {\hspace{-5pt} \tiny $\pm 0.49$}  & -21.33 {\hspace{-5pt} \tiny $\pm 0.78$} \\

GEM  &59.3 {\hspace{-5pt} \tiny $\pm 2.92$} & 66.57 {\hspace{-5pt} \tiny $\pm 2.03$} & -19.76 {\hspace{-5pt} \tiny $\pm 2.93$}  &56.14 {\hspace{-5pt} \tiny $\pm 1.42$} & 61.52 {\hspace{-5pt} \tiny $\pm 0.63$} & -25.56 {\hspace{-5pt} \tiny $\pm 2.2$}  &60.66 {\hspace{-5pt} \tiny $\pm 2.03$} & 63.17 {\hspace{-5pt} \tiny $\pm 0.83$} & -17.98 {\hspace{-5pt} \tiny $\pm 1.35$}  &61.5 {\hspace{-5pt} \tiny $\pm 1.01$} & 58.65 {\hspace{-5pt} \tiny $\pm 1.22$} & -18.27 {\hspace{-5pt} \tiny $\pm 2.08$}
&  59.4 {\hspace{-5pt} \tiny $\pm 0.72$}  & 62.48 {\hspace{-5pt} \tiny $\pm 0.54$}  & -20.39 {\hspace{-5pt} \tiny $\pm 0.56$} \\

AGEM  &62.77 {\hspace{-5pt} \tiny $\pm 2.35$} & 67.85 {\hspace{-5pt} \tiny $\pm 1.29$} & -15.47 {\hspace{-5pt} \tiny $\pm 1.83$}  &66.73 {\hspace{-5pt} \tiny $\pm 1.17$} & 68.25 {\hspace{-5pt} \tiny $\pm 1.03$} & -13.13 {\hspace{-5pt} \tiny $\pm 1.77$}  &64.8 {\hspace{-5pt} \tiny $\pm 2.45$} & 61.91 {\hspace{-5pt} \tiny $\pm 1.05$} &  \textcolor{red}{4.13} {\hspace{-5pt} \tiny $\pm 2.73$}  &64.63 {\hspace{-5pt} \tiny $\pm 2.78$} & 58.39 {\hspace{-5pt} \tiny $\pm 2.79$} & 4.36 {\hspace{-5pt} \tiny $\pm 4.26$}
&  64.73 {\hspace{-5pt} \tiny $\pm 0.61$}  & 64.1 {\hspace{-5pt} \tiny $\pm 0.73$}  &  \textcolor{red}{5.03} {\hspace{-5pt} \tiny $\pm 1.01$} \\

 LR  &67.93 {\hspace{-5pt} \tiny $\pm 1.9$} & \textcolor{blue}{79.11} {\hspace{-5pt} \tiny $\pm 0.76$} & \textcolor{blue}{-11.82} {\hspace{-5pt} \tiny $\pm 1.33$}  &78.9 {\hspace{-5pt} \tiny $\pm 1.96$} & 82.51 {\hspace{-5pt} \tiny $\pm 1.19$} & -9.09 {\hspace{-5pt} \tiny $\pm 1.51$}  &78.47 {\hspace{-5pt} \tiny $\pm 1.7$} & 77.01 {\hspace{-5pt} \tiny $\pm 0.72$} & -11.75 {\hspace{-5pt} \tiny $\pm 0.83$}  &77.33 {\hspace{-5pt} \tiny $\pm 2.72$} & 78.71 {\hspace{-5pt} \tiny $\pm 0.81$} & -11.62 {\hspace{-5pt} \tiny $\pm 3.05$}
&  75.66 {\hspace{-5pt} \tiny $\pm 0.39$}  & 79.33 {\hspace{-5pt} \tiny $\pm 0.19$}  & -11.07 {\hspace{-5pt} \tiny $\pm 0.83$} \\ \hline

LwF  &51.17 {\hspace{-5pt} \tiny $\pm 1.02$} & 59.88 {\hspace{-5pt} \tiny $\pm 1.86$} & -30.53 {\hspace{-5pt} \tiny $\pm 1.3$}  &65.4 {\hspace{-5pt} \tiny $\pm 0.86$} & 62.79 {\hspace{-5pt} \tiny $\pm 1.26$} & -24.49 {\hspace{-5pt} \tiny $\pm 1.37$}  &55.3 {\hspace{-5pt} \tiny $\pm 1.5$} & 59.85 {\hspace{-5pt} \tiny $\pm 0.9$} & -22.54 {\hspace{-5pt} \tiny $\pm 3.02$}  &60.67 {\hspace{-5pt} \tiny $\pm 2.11$} & 60.23 {\hspace{-5pt} \tiny $\pm 1.35$} & -16.56 {\hspace{-5pt} \tiny $\pm 4.43$}
&  58.14 {\hspace{-5pt} \tiny $\pm 0.49$}  & 60.69 {\hspace{-5pt} \tiny $\pm 0.34$}  & -23.53 {\hspace{-5pt} \tiny $\pm 1.3$} \\ 

EWC  &58.07 {\hspace{-5pt} \tiny $\pm 1.74$} & 69.66 {\hspace{-5pt} \tiny $\pm 0.8$} & -36.71 {\hspace{-5pt} \tiny $\pm 0.66$}  &67.7 {\hspace{-5pt} \tiny $\pm 0.75$} & 67.45 {\hspace{-5pt} \tiny $\pm 0.55$} & -34.82 {\hspace{-5pt} \tiny $\pm 1.1$}  &59.33 {\hspace{-5pt} \tiny $\pm 1.06$} & 64.24 {\hspace{-5pt} \tiny $\pm 1.1$} & -32.36 {\hspace{-5pt} \tiny $\pm 1.38$}  &62.07 {\hspace{-5pt} \tiny $\pm 2.66$} & 65.45 {\hspace{-5pt} \tiny $\pm 1.38$} & -33.27 {\hspace{-5pt} \tiny $\pm 3.97$}
&  61.79 {\hspace{-5pt} \tiny $\pm 0.73$}  & 66.7 {\hspace{-5pt} \tiny $\pm 0.31$}  & -34.29 {\hspace{-5pt} \tiny $\pm 1.29$} \\ 

SI  &58.97 {\hspace{-5pt} \tiny $\pm 3.28$} & 70.31 {\hspace{-5pt} \tiny $\pm 2.22$} & -34.73 {\hspace{-5pt} \tiny $\pm 4.4$}  &64.3 {\hspace{-5pt} \tiny $\pm 1.62$} & 65.95 {\hspace{-5pt} \tiny $\pm 0.67$} & -38.76 {\hspace{-5pt} \tiny $\pm 1.67$}  &57.37 {\hspace{-5pt} \tiny $\pm 2.59$} & 62.97 {\hspace{-5pt} \tiny $\pm 0.82$} & -32.55 {\hspace{-5pt} \tiny $\pm 2.1$}  &61.9 {\hspace{-5pt} \tiny $\pm 1.65$} & 65.63 {\hspace{-5pt} \tiny $\pm 1.14$} & -33.11 {\hspace{-5pt} \tiny $\pm 1.41$}
&  60.64 {\hspace{-5pt} \tiny $\pm 0.69$}  & 66.22 {\hspace{-5pt} \tiny $\pm 0.61$}  & -34.79 {\hspace{-5pt} \tiny $\pm 1.18$} \\

GLRCL
& \textcolor{red}{ 73.17} {\hspace{-5pt} \tiny $\pm 1.26$} &\textcolor{red}{ 82.65} {\hspace{-5pt} \tiny $\pm 1.04$} & \textcolor{red}{-6.29} {\hspace{-5pt} \tiny $\pm 2.18$}
 &  \textcolor{red}{82.63} {\hspace{-5pt} \tiny $\pm 1.08$} & \textcolor{red}{85.49} {\hspace{-5pt} \tiny $\pm 0.86$} & \textcolor{red}{-3.60 }{\hspace{-5pt} \tiny $\pm 1.19$}
 &  \textcolor{blue}{83.13} {\hspace{-5pt} \tiny $\pm 1.15$} & \textcolor{blue}{80.64} {\hspace{-5pt} \tiny $\pm 0.64$} & \textcolor{blue}{-5.02} {\hspace{-5pt} \tiny $\pm 1.67$}
 &  \textcolor{red}{83.30} {\hspace{-5pt} \tiny $\pm 1.05$} &\textcolor{red}{ 82.72} {\hspace{-5pt} \tiny $\pm 1.06$} & \textcolor{red}{-2.91} {\hspace{-5pt} \tiny $\pm 1.31$}
 &  \textcolor{red}{80.56} {\hspace{-5pt} \tiny $\pm 4.42$} & \textcolor{red}{82.88} {\hspace{-5pt} \tiny $\pm 1.95$} & \textcolor{red}{-4.46} {\hspace{-5pt} \tiny $\pm 2.09$}\\

Our
 &  \textcolor{blue}{68.73} {\hspace{-5pt} \tiny $\pm 1.54$} & 78.84 {\hspace{-5pt} \tiny $\pm 0.69$} & -11.84 {\hspace{-5pt} \tiny $\pm 0.36$}
 & \textcolor{blue}{ 79.40} {\hspace{-5pt} \tiny $\pm 1.36$} &\textcolor{blue}{ 82.70} {\hspace{-5pt} \tiny $\pm 0.95$} & \textcolor{blue}{-5.73 }{\hspace{-5pt} \tiny $\pm 1.60$}
 & \textcolor{red}{ 83.30 }{\hspace{-5pt} \tiny $\pm 1.31$} & \textcolor{red}{81.16} {\hspace{-5pt} \tiny $\pm 0.63$} & \textcolor{red}{-3.80} {\hspace{-5pt} \tiny $\pm 1.47$}
 &  \textcolor{blue}{80.20} {\hspace{-5pt} \tiny $\pm 0.66$} &\textcolor{blue}{ 80.93} {\hspace{-5pt} \tiny $\pm 1.18$} & \textcolor{blue}{-5.18 }{\hspace{-5pt} \tiny $\pm 1.54$}
 &  \textcolor{blue}{77.91} {\hspace{-5pt} \tiny $\pm 5.64$} & \textcolor{blue}{80.91} {\hspace{-5pt} \tiny $\pm 1.64$} & -6.64 {\hspace{-5pt} \tiny $\pm 3.37$}\\

 \hline
\end{tabular}%
}
\end{table*}

\begin{table*}[!t]
\centering
\caption{ Performance comparison under \textbf{RetCCL} backbone (\textcolor{red}{best result}, \textcolor{blue}{second best result} in continual learning methods)}
 \label{tab:comp_cl_all_RetCCL}
\resizebox{\textwidth}{!}{%
\def\arraystretch{0.9}
\setlength{\tabcolsep}{5pt}
\begin{tabular}{|l|ccc|ccc|ccc|ccc|ccc|}
\hline
\rowcolor{gray!10} 
 & \multicolumn{3}{c|}{\textbf{Seq.1}} & \multicolumn{3}{c|}{\textbf{Seq.2}} & \multicolumn{3}{c|}{\textbf{Seq.3}} & \multicolumn{3}{c|}{\textbf{Seq.4}} & \multicolumn{3}{c|}{\textbf{Average}} \\ \hline
\rowcolor{gray!10} 
\begin{tabular}[c]{@{}c@{}}\textbf{Method} \end{tabular} & 
\textbf{ACC}$\uparrow$ & \textbf{ILM}$\uparrow$ & \textbf{BWT}$\uparrow$ &
\textbf{ACC}$\uparrow$ & \textbf{ILM}$\uparrow$ & \textbf{BWT}$\uparrow$ &
\textbf{ACC}$\uparrow$ & \textbf{ILM}$\uparrow$ & \textbf{BWT}$\uparrow$ & 
\textbf{ACC}$\uparrow$ & \textbf{ILM}$\uparrow$ & \textbf{BWT}$\uparrow$ &
\textbf{ACC}$\uparrow$ & \textbf{ILM}$\uparrow$ & \textbf{BWT}$\uparrow$ \\
\hline

Naive  &66.23 {\hspace{-5pt} \tiny $\pm 2.21$} & 79.21 {\hspace{-5pt} \tiny $\pm 1.47$} & -26.09 {\hspace{-5pt} \tiny $\pm 2.32$}  &71.0 {\hspace{-5pt} \tiny $\pm 2.27$} & 74.15 {\hspace{-5pt} \tiny $\pm 1.05$} & -32.51 {\hspace{-5pt} \tiny $\pm 1.99$}  &58.3 {\hspace{-5pt} \tiny $\pm 1.3$} & 69.24 {\hspace{-5pt} \tiny $\pm 0.98$} & -31.15 {\hspace{-5pt} \tiny $\pm 2.28$}  &75.47 {\hspace{-5pt} \tiny $\pm 1.85$} & 74.95 {\hspace{-5pt} \tiny $\pm 0.88$} & -23.69 {\hspace{-5pt} \tiny $\pm 3.17$}
&  67.75 {\hspace{-5pt} \tiny $\pm 0.39$}  & 74.39 {\hspace{-5pt} \tiny $\pm 0.22$}  & -28.36 {\hspace{-5pt} \tiny $\pm 0.44$} \\ 

Joint
& 91.63 {\hspace{-5pt} \tiny $\pm 0.97$}
&-&-& 91.37 {\hspace{-5pt} \tiny $\pm 0.55$}
&-&-& 90.83 {\hspace{-5pt} \tiny $\pm 1.29$}
&-&-& 91.77 {\hspace{-5pt} \tiny $\pm 0.95$}
&-&-&91.40{\hspace{-5pt} \tiny $\pm 0.41$}
&-&-\\\hline

DER  &70.17 {\hspace{-5pt} \tiny $\pm 3.05$} & 80.92 {\hspace{-5pt} \tiny $\pm 2.2$} & -23.58 {\hspace{-5pt} \tiny $\pm 3.43$}  &70.47 {\hspace{-5pt} \tiny $\pm 2.42$} & 73.53 {\hspace{-5pt} \tiny $\pm 1.1$} & -34.33 {\hspace{-5pt} \tiny $\pm 2.66$}  &59.2 {\hspace{-5pt} \tiny $\pm 2.07$} & 69.32 {\hspace{-5pt} \tiny $\pm 1.13$} & -30.76 {\hspace{-5pt} \tiny $\pm 2.64$}  &74.73 {\hspace{-5pt} \tiny $\pm 4.48$} & 73.81 {\hspace{-5pt} \tiny $\pm 1.93$} & -22.82 {\hspace{-5pt} \tiny $\pm 2.91$}
&  68.64 {\hspace{-5pt} \tiny $\pm 0.92$}  & 74.4 {\hspace{-5pt} \tiny $\pm 0.48$}  & -27.87 {\hspace{-5pt} \tiny $\pm 0.32$} \\

 ER & \textcolor{red}{79.67} {\hspace{-5pt} \tiny $\pm 1.58$} &  \textcolor{red}{87.2} {\hspace{-5pt} \tiny $\pm 0.87$} & -12.2 {\hspace{-5pt} \tiny $\pm 1.14$}  &86.17 {\hspace{-5pt} \tiny $\pm 1.45$} & 86.71 {\hspace{-5pt} \tiny $\pm 1.11$} & -10.76 {\hspace{-5pt} \tiny $\pm 1.14$}  &77.03 {\hspace{-5pt} \tiny $\pm 2.84$} & 80.2 {\hspace{-5pt} \tiny $\pm 1.84$} & -12.16 {\hspace{-5pt} \tiny $\pm 1.69$}  &81.7 {\hspace{-5pt} \tiny $\pm 3.53$} & 83.43 {\hspace{-5pt} \tiny $\pm 1.99$} & -8.38 {\hspace{-5pt} \tiny $\pm 4.81$}
&  81.14 {\hspace{-5pt} \tiny $\pm 0.87$}  &  \textcolor{blue}{84.38} {\hspace{-5pt} \tiny $\pm 0.47$}  & -10.88 {\hspace{-5pt} \tiny $\pm 1.53$} \\

GEM  &72.37 {\hspace{-5pt} \tiny $\pm 2.54$} & 78.13 {\hspace{-5pt} \tiny $\pm 1.31$} &  \textcolor{blue}{-4.2} {\hspace{-5pt} \tiny $\pm 2.48$}  &65.23 {\hspace{-5pt} \tiny $\pm 2.57$} & 75.21 {\hspace{-5pt} \tiny $\pm 1.07$} & -21.29 {\hspace{-5pt} \tiny $\pm 1.45$}  &76.47 {\hspace{-5pt} \tiny $\pm 0.51$} & 75.39 {\hspace{-5pt} \tiny $\pm 1.3$} & -11.29 {\hspace{-5pt} \tiny $\pm 1.55$}  &72.8 {\hspace{-5pt} \tiny $\pm 3.05$} & 72.47 {\hspace{-5pt} \tiny $\pm 1.56$} & -18.53 {\hspace{-5pt} \tiny $\pm 2.43$}
&  71.72 {\hspace{-5pt} \tiny $\pm 0.98$}  & 75.3 {\hspace{-5pt} \tiny $\pm 0.17$}  & -13.83 {\hspace{-5pt} \tiny $\pm 0.48$} \\

AGEM  &62.0 {\hspace{-5pt} \tiny $\pm 3.55$} & 65.35 {\hspace{-5pt} \tiny $\pm 1.65$} &  \textcolor{red}{6.67} {\hspace{-5pt} \tiny $\pm 6.07$}  &73.33 {\hspace{-5pt} \tiny $\pm 4.32$} & 73.81 {\hspace{-5pt} \tiny $\pm 4.09$} & -5.42 {\hspace{-5pt} \tiny $\pm 1.38$}  &68.47 {\hspace{-5pt} \tiny $\pm 3.74$} & 63.65 {\hspace{-5pt} \tiny $\pm 3.63$} &  \textcolor{red}{12.04 }{\hspace{-5pt} \tiny $\pm 2.17$}  &73.94 {\hspace{-5pt} \tiny $\pm 1.23$} & 61.91 {\hspace{-5pt} \tiny $\pm 2.66$} &  \textcolor{red}{11.85} {\hspace{-5pt} \tiny $\pm 2.12$}
&  69.44 {\hspace{-5pt} \tiny $\pm 1.18$}  & 66.18 {\hspace{-5pt} \tiny $\pm 0.94$}  &  \textcolor{red}{6.28} {\hspace{-5pt} \tiny $\pm 1.84$} \\

 LR  &70.07 {\hspace{-5pt} \tiny $\pm 4.76$} & 81.13 {\hspace{-5pt} \tiny $\pm 2.43$} & -14.87 {\hspace{-5pt} \tiny $\pm 2.88$}  &86.93 {\hspace{-5pt} \tiny $\pm 0.94$} & 88.88 {\hspace{-5pt} \tiny $\pm 0.43$} & -6.11 {\hspace{-5pt} \tiny $\pm 1.22$}  & \textcolor{blue}{81.97} {\hspace{-5pt} \tiny $\pm 2.47$} & \textcolor{blue}{ 81.96 }{\hspace{-5pt} \tiny $\pm 1.65$} & -7.09 {\hspace{-5pt} \tiny $\pm 2.87$}  & \textcolor{red}{86.8} {\hspace{-5pt} \tiny $\pm 1.72$} &  \textcolor{red}{85.53 }{\hspace{-5pt} \tiny $\pm 1.11$} & -4.71 {\hspace{-5pt} \tiny $\pm 2.14$}
&   \textcolor{blue}{81.44} {\hspace{-5pt} \tiny $\pm 1.43$}  &  \textcolor{blue}{84.38} {\hspace{-5pt} \tiny $\pm 0.73$}  & -8.2 {\hspace{-5pt} \tiny $\pm 0.68$} \\ \hline

LwF  &64.63 {\hspace{-5pt} \tiny $\pm 2.66$} & 76.05 {\hspace{-5pt} \tiny $\pm 3.5$} & -15.31 {\hspace{-5pt} \tiny $\pm 1.26$}  &69.03 {\hspace{-5pt} \tiny $\pm 2.8$} & 69.59 {\hspace{-5pt} \tiny $\pm 1.31$} & -29.8 {\hspace{-5pt} \tiny $\pm 2.55$}  &61.23 {\hspace{-5pt} \tiny $\pm 0.73$} & 67.73 {\hspace{-5pt} \tiny $\pm 1.26$} & -18.42 {\hspace{-5pt} \tiny $\pm 2.64$}  &75.2 {\hspace{-5pt} \tiny $\pm 1.77$} & 69.82 {\hspace{-5pt} \tiny $\pm 1.24$} & -12.02 {\hspace{-5pt} \tiny $\pm 0.84$}
&  67.52 {\hspace{-5pt} \tiny $\pm 0.83$}  & 70.8 {\hspace{-5pt} \tiny $\pm 0.97$}  & -18.89 {\hspace{-5pt} \tiny $\pm 0.79$} \\

EWC  &68.83 {\hspace{-5pt} \tiny $\pm 4.3$} & 78.55 {\hspace{-5pt} \tiny $\pm 2.07$} & -26.76 {\hspace{-5pt} \tiny $\pm 4.18$}  &72.67 {\hspace{-5pt} \tiny $\pm 2.5$} & 74.58 {\hspace{-5pt} \tiny $\pm 2.09$} & -31.4 {\hspace{-5pt} \tiny $\pm 3.44$}  &58.43 {\hspace{-5pt} \tiny $\pm 1.03$} & 69.27 {\hspace{-5pt} \tiny $\pm 0.78$} & -28.78 {\hspace{-5pt} \tiny $\pm 1.99$}  &75.63 {\hspace{-5pt} \tiny $\pm 1.61$} & 74.9 {\hspace{-5pt} \tiny $\pm 0.73$} & -23.29 {\hspace{-5pt} \tiny $\pm 2.44$}
&  68.89 {\hspace{-5pt} \tiny $\pm 1.24$}  & 74.32 {\hspace{-5pt} \tiny $\pm 0.66$}  & -27.56 {\hspace{-5pt} \tiny $\pm 0.85$} \\ 

SI  &65.67 {\hspace{-5pt} \tiny $\pm 2.42$} & 76.65 {\hspace{-5pt} \tiny $\pm 1.18$} & -28.35 {\hspace{-5pt} \tiny $\pm 1.3$}  &74.73 {\hspace{-5pt} \tiny $\pm 3.16$} & 75.61 {\hspace{-5pt} \tiny $\pm 2.48$} & -29.42 {\hspace{-5pt} \tiny $\pm 3.23$}  &62.26 {\hspace{-5pt} \tiny $\pm 1.79$} & 71.05 {\hspace{-5pt} \tiny $\pm 0.68$} & -26.38 {\hspace{-5pt} \tiny $\pm 2.69$}  &76.7 {\hspace{-5pt} \tiny $\pm 4.95$} & 75.27 {\hspace{-5pt} \tiny $\pm 2.27$} & -22.69 {\hspace{-5pt} \tiny $\pm 5.03$}
&  69.84 {\hspace{-5pt} \tiny $\pm 1.18$}  & 74.64 {\hspace{-5pt} \tiny $\pm 0.75$}  & -26.71 {\hspace{-5pt} \tiny $\pm 1.34$} \\

GLRCL
 &  63.50 {\hspace{-5pt} \tiny $\pm 2.40$} & 77.16 {\hspace{-5pt} \tiny $\pm 1.33$} & -21.47 {\hspace{-5pt} \tiny $\pm 1.37$}
 &   \textcolor{red}{88.70} {\hspace{-5pt} \tiny $\pm 1.00$} &  \textcolor{red}{90.13} {\hspace{-5pt} \tiny $\pm 0.97$} &  \textcolor{blue}{-3.94} {\hspace{-5pt} \tiny $\pm 1.16$}
 &  80.97 {\hspace{-5pt} \tiny $\pm 1.38$} & 78.87 {\hspace{-5pt} \tiny $\pm 1.69$} & -15.00 {\hspace{-5pt} \tiny $\pm 3.39$}
 &  78.23 {\hspace{-5pt} \tiny $\pm 2.62$} & 79.52 {\hspace{-5pt} \tiny $\pm 1.79$} & -13.98 {\hspace{-5pt} \tiny $\pm 5.00$}
 &  77.85 {\hspace{-5pt} \tiny $\pm 9.34$} & 81.42 {\hspace{-5pt} \tiny $\pm 5.31$} & -13.60 {\hspace{-5pt} \tiny $\pm 7.02$}\\

Our
 &   \textcolor{blue}{78.23} {\hspace{-5pt} \tiny $\pm 2.45$} &  \textcolor{blue}{85.56} {\hspace{-5pt} \tiny $\pm 1.18$} &  -6.62 {\hspace{-5pt} \tiny $\pm 2.05$}
 &   \textcolor{blue}{87.73} {\hspace{-5pt} \tiny $\pm 0.88$} &  \textcolor{blue}{89.43} {\hspace{-5pt} \tiny $\pm 0.99$} &  \textcolor{red}{-3.33} {\hspace{-5pt} \tiny $\pm 1.32$}
 &   \textcolor{red}{84.00} {\hspace{-5pt} \tiny $\pm 2.21$} &  \textcolor{red}{82.72} {\hspace{-5pt} \tiny $\pm 2.02$} &  \textcolor{blue}{-5.36} {\hspace{-5pt} \tiny $\pm 2.70$}
 &  \textcolor{blue}{ 85.83} {\hspace{-5pt} \tiny $\pm 2.03$} &  \textcolor{blue}{85.12} {\hspace{-5pt} \tiny $\pm 1.98$} &  \textcolor{blue}{-3.53 }{\hspace{-5pt} \tiny $\pm 1.55$}
 &  \textcolor{red}{ 83.95} {\hspace{-5pt} \tiny $\pm 4.07$} &  \textcolor{red}{85.71} {\hspace{-5pt} \tiny $\pm 2.89$} &  \textcolor{blue}{-4.71} {\hspace{-5pt} \tiny $\pm 2.40$}\\

  \hline
\end{tabular}%
}
\end{table*}

\begin{table*}[!t]
\centering
\caption{ Performance comparison under \textbf{CTransPath} backbone (\textcolor{red}{best result}, \textcolor{blue}{second best result} in continual learning methods)}
 \label{tab:comp_cl_all_CTransPath}
\resizebox{\textwidth}{!}{%
\def\arraystretch{0.9}
\setlength{\tabcolsep}{5pt}
\begin{tabular}{|l|ccc|ccc|ccc|ccc|ccc|}
\hline
\rowcolor{gray!10} 
 & \multicolumn{3}{c|}{\textbf{Seq.1}} & \multicolumn{3}{c|}{\textbf{Seq.2}} & \multicolumn{3}{c|}{\textbf{Seq.3}} & \multicolumn{3}{c|}{\textbf{Seq.4}} & \multicolumn{3}{c|}{\textbf{Average}} \\ \hline
\rowcolor{gray!10} 
\begin{tabular}[c]{@{}c@{}}\textbf{Method} \end{tabular} & \textbf{ACC}$\uparrow$ & \textbf{ILM}$\uparrow$ & \textbf{BWT}$\uparrow$ &
\textbf{ACC}$\uparrow$ & \textbf{ILM}$\uparrow$ & \textbf{BWT}$\uparrow$ &
\textbf{ACC}$\uparrow$ & \textbf{ILM}$\uparrow$ & \textbf{BWT}$\uparrow$ & 
\textbf{ACC}$\uparrow$ & \textbf{ILM}$\uparrow$ & \textbf{BWT}$\uparrow$ &
\textbf{ACC}$\uparrow$ & \textbf{ILM}$\uparrow$ & \textbf{BWT}$\uparrow$ \\

Naive  &57.47 {\hspace{-5pt} \tiny $\pm 3.37$} & 73.76 {\hspace{-5pt} \tiny $\pm 1.58$} & -35.51 {\hspace{-5pt} \tiny $\pm 2.67$}  &78.27 {\hspace{-5pt} \tiny $\pm 1.62$} & 79.93 {\hspace{-5pt} \tiny $\pm 1.54$} & -21.73 {\hspace{-5pt} \tiny $\pm 2.32$}  &71.93 {\hspace{-5pt} \tiny $\pm 1.66$} & 72.81 {\hspace{-5pt} \tiny $\pm 0.73$} & -26.56 {\hspace{-5pt} \tiny $\pm 1.2$}  &66.63 {\hspace{-5pt} \tiny $\pm 1.9$} & 72.21 {\hspace{-5pt} \tiny $\pm 0.82$} & -32.27 {\hspace{-5pt} \tiny $\pm 2.45$}
&  68.58 {\hspace{-5pt} \tiny $\pm 0.72$}  & 74.68 {\hspace{-5pt} \tiny $\pm 0.39$}  & -29.02 {\hspace{-5pt} \tiny $\pm 0.57$} \\

Joint
& 91.93 {\hspace{-5pt} \tiny $\pm 1.06$}
&-&-& 92.33 {\hspace{-5pt} \tiny $\pm 0.59$}
&-&-& 92.27 {\hspace{-5pt} \tiny $\pm 0.83$}
&-&-& 91.77 {\hspace{-5pt} \tiny $\pm 1.15$}
&-&-&92.07{\hspace{-5pt} \tiny $\pm 0.27$}&-&-\\\hline

DER  &58.0 {\hspace{-5pt} \tiny $\pm 4.2$} & 73.83 {\hspace{-5pt} \tiny $\pm 2.09$} & -35.45 {\hspace{-5pt} \tiny $\pm 3.05$}  &78.86 {\hspace{-5pt} \tiny $\pm 2.52$} & 79.28 {\hspace{-5pt} \tiny $\pm 1.37$} & -21.73 {\hspace{-5pt} \tiny $\pm 1.52$}  &72.9 {\hspace{-5pt} \tiny $\pm 1.78$} & 73.45 {\hspace{-5pt} \tiny $\pm 0.92$} & -26.47 {\hspace{-5pt} \tiny $\pm 2.62$}  &64.2 {\hspace{-5pt} \tiny $\pm 1.88$} & 70.61 {\hspace{-5pt} \tiny $\pm 0.59$} & -33.05 {\hspace{-5pt} \tiny $\pm 2.42$}
&  68.49 {\hspace{-5pt} \tiny $\pm 0.97$}  & 74.29 {\hspace{-5pt} \tiny $\pm 0.56$}  & -29.18 {\hspace{-5pt} \tiny $\pm 0.56$} \\

 ER &83.13 {\hspace{-5pt} \tiny $\pm 3.12$} & 90.07 {\hspace{-5pt} \tiny $\pm 1.81$} & -7.89 {\hspace{-5pt} \tiny $\pm 3.11$}  &82.56 {\hspace{-5pt} \tiny $\pm 3.25$} & 85.59 {\hspace{-5pt} \tiny $\pm 1.48$} & -11.16 {\hspace{-5pt} \tiny $\pm 3.02$}  &80.13 {\hspace{-5pt} \tiny $\pm 2.49$} & 80.77 {\hspace{-5pt} \tiny $\pm 1.83$} & -15.73 {\hspace{-5pt} \tiny $\pm 1.26$}  &85.0 {\hspace{-5pt} \tiny $\pm 2.76$} & 84.15 {\hspace{-5pt} \tiny $\pm 1.25$} & -11.31 {\hspace{-5pt} \tiny $\pm 2.66$}
&  82.7 {\hspace{-5pt} \tiny $\pm 0.3$}  & 85.15 {\hspace{-5pt} \tiny $\pm 0.24$}  & -11.52 {\hspace{-5pt} \tiny $\pm 0.74$} \\

GEM  &79.6 {\hspace{-5pt} \tiny $\pm 2.82$} & 87.17 {\hspace{-5pt} \tiny $\pm 1.59$} & -8.35 {\hspace{-5pt} \tiny $\pm 1.28$}  &81.17 {\hspace{-5pt} \tiny $\pm 1.51$} & 82.43 {\hspace{-5pt} \tiny $\pm 1.71$} & -16.87 {\hspace{-5pt} \tiny $\pm 3.77$}  &80.5 {\hspace{-5pt} \tiny $\pm 2.54$} & 79.51 {\hspace{-5pt} \tiny $\pm 1.54$} & -15.85 {\hspace{-5pt} \tiny $\pm 3.43$}  &80.47 {\hspace{-5pt} \tiny $\pm 3.92$} & 79.83 {\hspace{-5pt} \tiny $\pm 2.1$} & -16.09 {\hspace{-5pt} \tiny $\pm 3.34$}
&  80.44 {\hspace{-5pt} \tiny $\pm 0.86$}  & 82.24 {\hspace{-5pt} \tiny $\pm 0.22$}  & -14.29 {\hspace{-5pt} \tiny $\pm 0.98$} \\

AGEM  &82.23 {\hspace{-5pt} \tiny $\pm 1.33$} & 88.29 {\hspace{-5pt} \tiny $\pm 0.69$} & -5.29 {\hspace{-5pt} \tiny $\pm 1.16$}  &83.67 {\hspace{-5pt} \tiny $\pm 2.55$} & 83.64 {\hspace{-5pt} \tiny $\pm 1.87$} & -7.31 {\hspace{-5pt} \tiny $\pm 2.66$}  &81.97 {\hspace{-5pt} \tiny $\pm 1.07$} & 75.19 {\hspace{-5pt} \tiny $\pm 0.5$} &\textcolor{red}{ 9.25} {\hspace{-5pt} \tiny $\pm 2.26$}  &83.13 {\hspace{-5pt} \tiny $\pm 2.39$} & 75.39 {\hspace{-5pt} \tiny $\pm 2.05$} & \textcolor{red}{9.42} {\hspace{-5pt} \tiny $\pm 3.33$}
&  82.75 {\hspace{-5pt} \tiny $\pm 0.64$}  & 80.63 {\hspace{-5pt} \tiny $\pm 0.69$}  & \textcolor{red}{1.52} {\hspace{-5pt} \tiny $\pm 0.79$} \\

 LR  &84.27 {\hspace{-5pt} \tiny $\pm 1.68$} & 89.97 {\hspace{-5pt} \tiny $\pm 0.97$} & -5.64 {\hspace{-5pt} \tiny $\pm 2.44$}  &84.63 {\hspace{-5pt} \tiny $\pm 2.38$} & 87.03 {\hspace{-5pt} \tiny $\pm 1.42$} & -8.16 {\hspace{-5pt} \tiny $\pm 1.61$}  &81.43 {\hspace{-5pt} \tiny $\pm 1.89$} & 81.44 {\hspace{-5pt} \tiny $\pm 1.07$} & -11.4 {\hspace{-5pt} \tiny $\pm 3.98$}  &84.73 {\hspace{-5pt} \tiny $\pm 3.14$} & 84.53 {\hspace{-5pt} \tiny $\pm 2.02$} & -11.55 {\hspace{-5pt} \tiny $\pm 1.65$}
&  83.76 {\hspace{-5pt} \tiny $\pm 0.56$}  & 85.74 {\hspace{-5pt} \tiny $\pm 0.41$}  & -9.19 {\hspace{-5pt} \tiny $\pm 0.96$} \\ \hline

LwF  &57.47 {\hspace{-5pt} \tiny $\pm 2.14$} & 74.19 {\hspace{-5pt} \tiny $\pm 1.72$} & -21.87 {\hspace{-5pt} \tiny $\pm 3.46$}  &77.67 {\hspace{-5pt} \tiny $\pm 2.0$} & 68.12 {\hspace{-5pt} \tiny $\pm 1.84$} & -13.31 {\hspace{-5pt} \tiny $\pm 4.57$}  &76.96 {\hspace{-5pt} \tiny $\pm 2.5$} & 75.03 {\hspace{-5pt} \tiny $\pm 2.14$} & -5.44 {\hspace{-5pt} \tiny $\pm 3.88$}  &67.1 {\hspace{-5pt} \tiny $\pm 1.73$} & 71.53 {\hspace{-5pt} \tiny $\pm 1.69$} & -14.53 {\hspace{-5pt} \tiny $\pm 3.51$}
&  69.8 {\hspace{-5pt} \tiny $\pm 0.28$}  & 72.22 {\hspace{-5pt} \tiny $\pm 0.18$}  & -13.79 {\hspace{-5pt} \tiny $\pm 0.44$} \\ 

EWC  &57.77 {\hspace{-5pt} \tiny $\pm 2.78$} & 72.61 {\hspace{-5pt} \tiny $\pm 1.5$} & -36.64 {\hspace{-5pt} \tiny $\pm 2.69$}  &78.07 {\hspace{-5pt} \tiny $\pm 0.92$} & 79.23 {\hspace{-5pt} \tiny $\pm 2.03$} & -21.4 {\hspace{-5pt} \tiny $\pm 2.25$}  &71.03 {\hspace{-5pt} \tiny $\pm 1.79$} & 72.6 {\hspace{-5pt} \tiny $\pm 0.74$} & -25.96 {\hspace{-5pt} \tiny $\pm 3.72$}  &67.83 {\hspace{-5pt} \tiny $\pm 2.7$} & 72.6 {\hspace{-5pt} \tiny $\pm 1.52$} & -28.84 {\hspace{-5pt} \tiny $\pm 1.38$}
&  68.68 {\hspace{-5pt} \tiny $\pm 0.76$}  & 74.26 {\hspace{-5pt} \tiny $\pm 0.46$}  & -28.21 {\hspace{-5pt} \tiny $\pm 0.84$} \\

SI  &66.13 {\hspace{-5pt} \tiny $\pm 4.33$} & 80.36 {\hspace{-5pt} \tiny $\pm 2.62$} & -23.45 {\hspace{-5pt} \tiny $\pm 4.51$}  &79.27 {\hspace{-5pt} \tiny $\pm 2.29$} & 80.19 {\hspace{-5pt} \tiny $\pm 1.66$} & -20.53 {\hspace{-5pt} \tiny $\pm 2.28$}  &70.2 {\hspace{-5pt} \tiny $\pm 0.61$} & 71.71 {\hspace{-5pt} \tiny $\pm 0.9$} & -29.93 {\hspace{-5pt} \tiny $\pm 1.3$}  &71.03 {\hspace{-5pt} \tiny $\pm 3.77$} & 74.88 {\hspace{-5pt} \tiny $\pm 1.54$} & -26.91 {\hspace{-5pt} \tiny $\pm 2.42$}
&  71.66 {\hspace{-5pt} \tiny $\pm 1.44$}  & 76.78 {\hspace{-5pt} \tiny $\pm 0.61$}  & -25.2 {\hspace{-5pt} \tiny $\pm 1.17$} \\

GLRCL
 & \textcolor{blue}{ 87.97} {\hspace{-5pt} \tiny $\pm 1.18$} &\textcolor{blue}{ 93.19} {\hspace{-5pt} \tiny $\pm 0.57$} &\textcolor{blue}{ -2.89} {\hspace{-5pt} \tiny $\pm 1.10$}
 & \textcolor{blue}{ 87.37 }{\hspace{-5pt} \tiny $\pm 1.33$} &\textcolor{blue}{ 88.89} {\hspace{-5pt} \tiny $\pm 0.67$} & \textcolor{blue}{-4.84 }{\hspace{-5pt} \tiny $\pm 1.26$}
 & \textcolor{blue}{ 89.16} {\hspace{-5pt} \tiny $\pm 1.42$} & \textcolor{blue}{87.65} {\hspace{-5pt} \tiny $\pm 1.36$} & -4.40 {\hspace{-5pt} \tiny $\pm 1.02$}
 & \textcolor{blue}{ 89.00} {\hspace{-5pt} \tiny $\pm 1.82$} & \textcolor{blue}{88.55} {\hspace{-5pt} \tiny $\pm 1.17$} & -3.96 {\hspace{-5pt} \tiny $\pm 1.44$}
 & \textcolor{blue}{ 88.37} {\hspace{-5pt} \tiny $\pm 1.63$} &\textcolor{blue}{ 89.57} {\hspace{-5pt} \tiny $\pm 2.36$} & -4.02 {\hspace{-5pt} \tiny $\pm 1.42$}\\

Our
 & \textcolor{red}{ 88.20} {\hspace{-5pt} \tiny $\pm 1.81$} & \textcolor{red}{93.37} {\hspace{-5pt} \tiny $\pm 1.13$} & \textcolor{red}{-1.58 }{\hspace{-5pt} \tiny $\pm 1.13$}
 &  \textcolor{red}{90.43} {\hspace{-5pt} \tiny $\pm 1.22$} & \textcolor{red}{90.48} {\hspace{-5pt} \tiny $\pm 1.90$} & \textcolor{red}{-1.95} {\hspace{-5pt} \tiny $\pm 0.95$}
 &  \textcolor{red}{89.83 }{\hspace{-5pt} \tiny $\pm 1.14$} &\textcolor{red}{ 87.48 }{\hspace{-5pt} \tiny $\pm 1.53$} & \textcolor{blue}{-2.62} {\hspace{-5pt} \tiny $\pm 1.34$}
 & \textcolor{red}{ 89.73} {\hspace{-5pt} \tiny $\pm 1.91$} & \textcolor{red}{88.95} {\hspace{-5pt} \tiny $\pm 2.52$} & \textcolor{blue}{-4.27} {\hspace{-5pt} \tiny $\pm 2.48$}
 &  \textcolor{red}{89.55} {\hspace{-5pt} \tiny $\pm 1.76$} & \textcolor{red}{90.07 }{\hspace{-5pt} \tiny $\pm 2.85$} & \textcolor{blue}{-2.61} {\hspace{-5pt} \tiny $\pm 1.89$}\\

 \hline
\end{tabular}%
}
\end{table*}


\begin{table*}[!t]
\centering
\caption{ Performance comparison under \textbf{UNI} backbone (\textcolor{red}{best result}, \textcolor{blue}{second best result} in continual learning methods)}
 \label{tab:comp_cl_all_UNI}
\resizebox{\textwidth}{!}{%
\def\arraystretch{0.9}
\setlength{\tabcolsep}{5pt}
\begin{tabular}{|l|ccc|ccc|ccc|ccc|ccc|}
\hline
\rowcolor{gray!10} 
 & \multicolumn{3}{c|}{\textbf{Seq.1}} & \multicolumn{3}{c|}{\textbf{Seq.2}} & \multicolumn{3}{c|}{\textbf{Seq.3}} & \multicolumn{3}{c|}{\textbf{Seq.4}} & \multicolumn{3}{c|}{\textbf{Average}} \\ \hline
\rowcolor{gray!10} 
\begin{tabular}[c]{@{}c@{}}\textbf{Method} \end{tabular} & \textbf{ACC}$\uparrow$ & \textbf{ILM}$\uparrow$ & \textbf{BWT}$\uparrow$ &
\textbf{ACC}$\uparrow$ & \textbf{ILM}$\uparrow$ & \textbf{BWT}$\uparrow$ &
\textbf{ACC}$\uparrow$ & \textbf{ILM}$\uparrow$ & \textbf{BWT}$\uparrow$ & 
\textbf{ACC}$\uparrow$ & \textbf{ILM}$\uparrow$ & \textbf{BWT}$\uparrow$ &
\textbf{ACC}$\uparrow$ & \textbf{ILM}$\uparrow$ & \textbf{BWT}$\uparrow$ \\
\hline

Naive  &78.3 {\hspace{-5pt} \tiny $\pm 5.47$} & 85.69 {\hspace{-5pt} \tiny $\pm 2.39$} & -15.98 {\hspace{-5pt} \tiny $\pm 4.12$}  &86.83 {\hspace{-5pt} \tiny $\pm 1.27$} & 87.87 {\hspace{-5pt} \tiny $\pm 0.6$} & -10.18 {\hspace{-5pt} \tiny $\pm 1.58$}  &79.67 {\hspace{-5pt} \tiny $\pm 1.61$} & 78.79 {\hspace{-5pt} \tiny $\pm 0.1$} & -20.15 {\hspace{-5pt} \tiny $\pm 1.08$}  &76.43 {\hspace{-5pt} \tiny $\pm 2.48$} & 77.87 {\hspace{-5pt} \tiny $\pm 0.88$} & -22.04 {\hspace{-5pt} \tiny $\pm 2.18$}
&  80.31 {\hspace{-5pt} \tiny $\pm 1.65$}  & 82.56 {\hspace{-5pt} \tiny $\pm 0.85$}  & -17.09 {\hspace{-5pt} \tiny $\pm 1.15$} \\ 

Joint
& 93.03 {\hspace{-5pt} \tiny $\pm 0.58$}
&-&-& 93.17 {\hspace{-5pt} \tiny $\pm 0.70$}
&-&-& 92.67 {\hspace{-5pt} \tiny $\pm 1.02$}
&-&-& 92.50 {\hspace{-5pt} \tiny $\pm 0.71$}
&-&-&92.82{\hspace{-5pt} \tiny $\pm 0.33$}&-&-\\\hline

DER  &75.43 {\hspace{-5pt} \tiny $\pm 1.34$} & 85.36 {\hspace{-5pt} \tiny $\pm 0.83$} & -17.34 {\hspace{-5pt} \tiny $\pm 1.88$}  &87.2 {\hspace{-5pt} \tiny $\pm 1.11$} & 88.29 {\hspace{-5pt} \tiny $\pm 0.89$} & -9.0 {\hspace{-5pt} \tiny $\pm 1.5$}  &82.07 {\hspace{-5pt} \tiny $\pm 1.74$} & 79.18 {\hspace{-5pt} \tiny $\pm 0.8$} & -19.07 {\hspace{-5pt} \tiny $\pm 0.65$}  &77.37 {\hspace{-5pt} \tiny $\pm 0.97$} & 78.44 {\hspace{-5pt} \tiny $\pm 0.71$} & -22.8 {\hspace{-5pt} \tiny $\pm 1.62$}
&  80.52 {\hspace{-5pt} \tiny $\pm 0.29$}  & 82.82 {\hspace{-5pt} \tiny $\pm 0.06$}  & -17.05 {\hspace{-5pt} \tiny $\pm 0.46$} \\

 ER &84.8 {\hspace{-5pt} \tiny $\pm 2.91$} & 90.81 {\hspace{-5pt} \tiny $\pm 2.04$} & -7.91 {\hspace{-5pt} \tiny $\pm 2.51$}  &88.1 {\hspace{-5pt} \tiny $\pm 1.98$} & 90.29 {\hspace{-5pt} \tiny $\pm 0.77$} & -5.4 {\hspace{-5pt} \tiny $\pm 0.76$}  &84.23 {\hspace{-5pt} \tiny $\pm 2.41$} & 84.41 {\hspace{-5pt} \tiny $\pm 0.41$} & -11.13 {\hspace{-5pt} \tiny $\pm 1.34$}  &85.77 {\hspace{-5pt} \tiny $\pm 2.06$} & 85.85 {\hspace{-5pt} \tiny $\pm 0.64$} & -10.6 {\hspace{-5pt} \tiny $\pm 1.9$}
&  85.72 {\hspace{-5pt} \tiny $\pm 0.37$}  & 87.84 {\hspace{-5pt} \tiny $\pm 0.63$}  & -8.76 {\hspace{-5pt} \tiny $\pm 0.65$} \\

GEM  &82.78 {\hspace{-5pt} \tiny $\pm 1.72$} & 88.49 {\hspace{-5pt} \tiny $\pm 0.94$} & -12.07 {\hspace{-5pt} \tiny $\pm 1.61$} &88.17 {\hspace{-5pt} \tiny $\pm 2.87$} & 89.19 {\hspace{-5pt} \tiny $\pm 1.66$} & -8.18 {\hspace{-5pt} \tiny $\pm 4.07$} &77.63 {\hspace{-5pt} \tiny $\pm 8.59$} & 80.61 {\hspace{-5pt} \tiny $\pm 3.6$} & -17.93 {\hspace{-5pt} \tiny $\pm 7.36$} &82.2 {\hspace{-5pt} \tiny $\pm 2.29$} & 81.51 {\hspace{-5pt} \tiny $\pm 1.72$} & -17.36 {\hspace{-5pt} \tiny $\pm 2.39$}
 &  82.7 {\hspace{-5pt} \tiny $\pm 2.76$}  & 84.95 {\hspace{-5pt} \tiny $\pm 0.98$}  & -13.88 {\hspace{-5pt} \tiny $\pm 2.21$} \\

AGEM  &63.9 {\hspace{-5pt} \tiny $\pm 25.13$} & 75.01 {\hspace{-5pt} \tiny $\pm 18.86$} & -23.22 {\hspace{-5pt} \tiny $\pm 18.09$}  &87.1 {\hspace{-5pt} \tiny $\pm 4.32$} & 89.07 {\hspace{-5pt} \tiny $\pm 2.43$} & -7.07 {\hspace{-5pt} \tiny $\pm 4.08$}  &82.77 {\hspace{-5pt} \tiny $\pm 4.73$} & 81.0 {\hspace{-5pt} \tiny $\pm 2.5$} & -7.24 {\hspace{-5pt} \tiny $\pm 4.42$}  &84.1 {\hspace{-5pt} \tiny $\pm 3.57$} & 80.67 {\hspace{-5pt} \tiny $\pm 1.78$} & -7.98 {\hspace{-5pt} \tiny $\pm 3.85$}
&  79.47 {\hspace{-5pt} \tiny $\pm 9.07$}  & 81.44 {\hspace{-5pt} \tiny $\pm 7.2$}  & -11.38 {\hspace{-5pt} \tiny $\pm 6.05$} \\

 LR  &88.23 {\hspace{-5pt} \tiny $\pm 0.83$} & 93.09 {\hspace{-5pt} \tiny $\pm 0.71$} & -4.27 {\hspace{-5pt} \tiny $\pm 1.0$}  &88.63 {\hspace{-5pt} \tiny $\pm 1.05$} & 90.17 {\hspace{-5pt} \tiny $\pm 0.93$} & -4.42 {\hspace{-5pt} \tiny $\pm 0.97$}  &82.43 {\hspace{-5pt} \tiny $\pm 3.33$} & 83.67 {\hspace{-5pt} \tiny $\pm 2.01$} & -11.95 {\hspace{-5pt} \tiny $\pm 2.83$}  &85.9 {\hspace{-5pt} \tiny $\pm 1.19$} & 86.77 {\hspace{-5pt} \tiny $\pm 0.97$} & -8.18 {\hspace{-5pt} \tiny $\pm 1.67$}
&  86.3 {\hspace{-5pt} \tiny $\pm 1.01$}  & 88.42 {\hspace{-5pt} \tiny $\pm 0.5$}  & -7.2 {\hspace{-5pt} \tiny $\pm 0.75$} \\ \hline

LwF  &81.7 {\hspace{-5pt} \tiny $\pm 2.02$} & 86.36 {\hspace{-5pt} \tiny $\pm 1.76$} & \textcolor{red}{0.09} {\hspace{-5pt} \tiny $\pm 2.24$}  &82.73 {\hspace{-5pt} \tiny $\pm 2.63$} & 83.05 {\hspace{-5pt} \tiny $\pm 4.07$} &  \textcolor{red}{-0.85} {\hspace{-5pt} \tiny $\pm 0.74$}  &84.5 {\hspace{-5pt} \tiny $\pm 1.84$} & 78.35 {\hspace{-5pt} \tiny $\pm 1.16$} &  \textcolor{red}{6.0 }{\hspace{-5pt} \tiny $\pm 1.76$}  &82.3 {\hspace{-5pt} \tiny $\pm 2.37$} & 77.6 {\hspace{-5pt} \tiny $\pm 2.93$} & \textcolor{red}{ 1.44 }{\hspace{-5pt} \tiny $\pm 2.04$}
&  82.81 {\hspace{-5pt} \tiny $\pm 0.31$}  & 81.34 {\hspace{-5pt} \tiny $\pm 1.12$}  &  \textcolor{red}{1.67} {\hspace{-5pt} \tiny $\pm 0.58$} \\

EWC  &79.0 {\hspace{-5pt} \tiny $\pm 5.51$} & 86.95 {\hspace{-5pt} \tiny $\pm 2.73$} & -14.02 {\hspace{-5pt} \tiny $\pm 5.0$}  &87.4 {\hspace{-5pt} \tiny $\pm 1.18$} & 87.84 {\hspace{-5pt} \tiny $\pm 0.39$} & -8.8 {\hspace{-5pt} \tiny $\pm 1.49$}  &81.97 {\hspace{-5pt} \tiny $\pm 1.95$} & 79.95 {\hspace{-5pt} \tiny $\pm 0.69$} & -18.51 {\hspace{-5pt} \tiny $\pm 1.52$}  &75.27 {\hspace{-5pt} \tiny $\pm 2.11$} & 77.62 {\hspace{-5pt} \tiny $\pm 0.82$} & -24.2 {\hspace{-5pt} \tiny $\pm 0.97$}
&  80.91 {\hspace{-5pt} \tiny $\pm 1.67$}  & 83.09 {\hspace{-5pt} \tiny $\pm 0.92$}  & -16.38 {\hspace{-5pt} \tiny $\pm 1.61$} \\

SI  &78.57 {\hspace{-5pt} \tiny $\pm 1.69$} & 86.68 {\hspace{-5pt} \tiny $\pm 1.4$} & -14.29 {\hspace{-5pt} \tiny $\pm 1.3$}  &84.87 {\hspace{-5pt} \tiny $\pm 2.29$} & 85.45 {\hspace{-5pt} \tiny $\pm 1.98$} & -12.64 {\hspace{-5pt} \tiny $\pm 2.17$}  &76.77 {\hspace{-5pt} \tiny $\pm 1.83$} & 78.65 {\hspace{-5pt} \tiny $\pm 0.79$} & -20.67 {\hspace{-5pt} \tiny $\pm 1.02$}  &74.2 {\hspace{-5pt} \tiny $\pm 5.82$} & 77.63 {\hspace{-5pt} \tiny $\pm 2.48$} & -24.04 {\hspace{-5pt} \tiny $\pm 3.51$}
&  78.6 {\hspace{-5pt} \tiny $\pm 1.7$}  & 82.1 {\hspace{-5pt} \tiny $\pm 0.63$}  & -17.91 {\hspace{-5pt} \tiny $\pm 0.97$} \\ 

 GLRCL
 &  \textcolor{blue}{ 90.13 }{\hspace{-5pt} \tiny $\pm 1.13$} &  \textcolor{blue}{93.93} {\hspace{-5pt} \tiny $\pm 0.83$} &  -2.36 {\hspace{-5pt} \tiny $\pm 1.06$}
 &   \textcolor{blue}{90.43} {\hspace{-5pt} \tiny $\pm 1.22$} &  \textcolor{blue}{91.18} {\hspace{-5pt} \tiny $\pm 1.26$} & -3.42 {\hspace{-5pt} \tiny $\pm 1.22$}
 &   \textcolor{blue}{89.90} {\hspace{-5pt} \tiny $\pm 0.49$} & \textcolor{blue}{ 88.28} {\hspace{-5pt} \tiny $\pm 0.22$} & -3.06 {\hspace{-5pt} \tiny $\pm 0.92$}
 &   \textcolor{blue}{88.27} {\hspace{-5pt} \tiny $\pm 1.06$} & \textcolor{blue}{ 87.39} {\hspace{-5pt} \tiny $\pm 1.77$} & -5.69 {\hspace{-5pt} \tiny $\pm 1.89$}
 &   \textcolor{blue}{89.68} {\hspace{-5pt} \tiny $\pm 1.32$} &  \textcolor{blue}{90.19 }{\hspace{-5pt} \tiny $\pm 2.83$} &  -3.63 {\hspace{-5pt} \tiny $\pm 1.82$}\\
 
Our
 & \textcolor{red}{ 91.33} {\hspace{-5pt} \tiny $\pm 0.63$} & \textcolor{red}{ 94.75} {\hspace{-5pt} \tiny $\pm 0.38$} &  \textcolor{blue}{-1.62} {\hspace{-5pt} \tiny $\pm 0.64$}
 &  \textcolor{red}{ 91.07} {\hspace{-5pt} \tiny $\pm 0.85$} &  \textcolor{red}{91.83} {\hspace{-5pt} \tiny $\pm 0.71$} &  \textcolor{blue}{-2.82} {\hspace{-5pt} \tiny $\pm 0.91$}
 &   \textcolor{red}{89.87} {\hspace{-5pt} \tiny $\pm 0.71$} &  \textcolor{red}{88.40} {\hspace{-5pt} \tiny $\pm 0.69$} & \textcolor{blue}{ -4.02} {\hspace{-5pt} \tiny $\pm 0.64$}
 &  \textcolor{red}{ 89.70} {\hspace{-5pt} \tiny $\pm 1.25$} &  \textcolor{red}{89.38} {\hspace{-5pt} \tiny $\pm 1.18$} & \textcolor{blue}{ -3.27} {\hspace{-5pt} \tiny $\pm 0.86$}
 &  \textcolor{red}{ 90.49} {\hspace{-5pt} \tiny $\pm 1.15$} &  \textcolor{red}{91.09 }{\hspace{-5pt} \tiny $\pm 2.58$} &  \textcolor{blue}{-2.93} {\hspace{-5pt} \tiny $\pm 1.16$}\\

\hline
\end{tabular}%
}
\end{table*}

\begin{figure*}[!htbp]
\centering
    \includegraphics[width=0.47\textwidth]{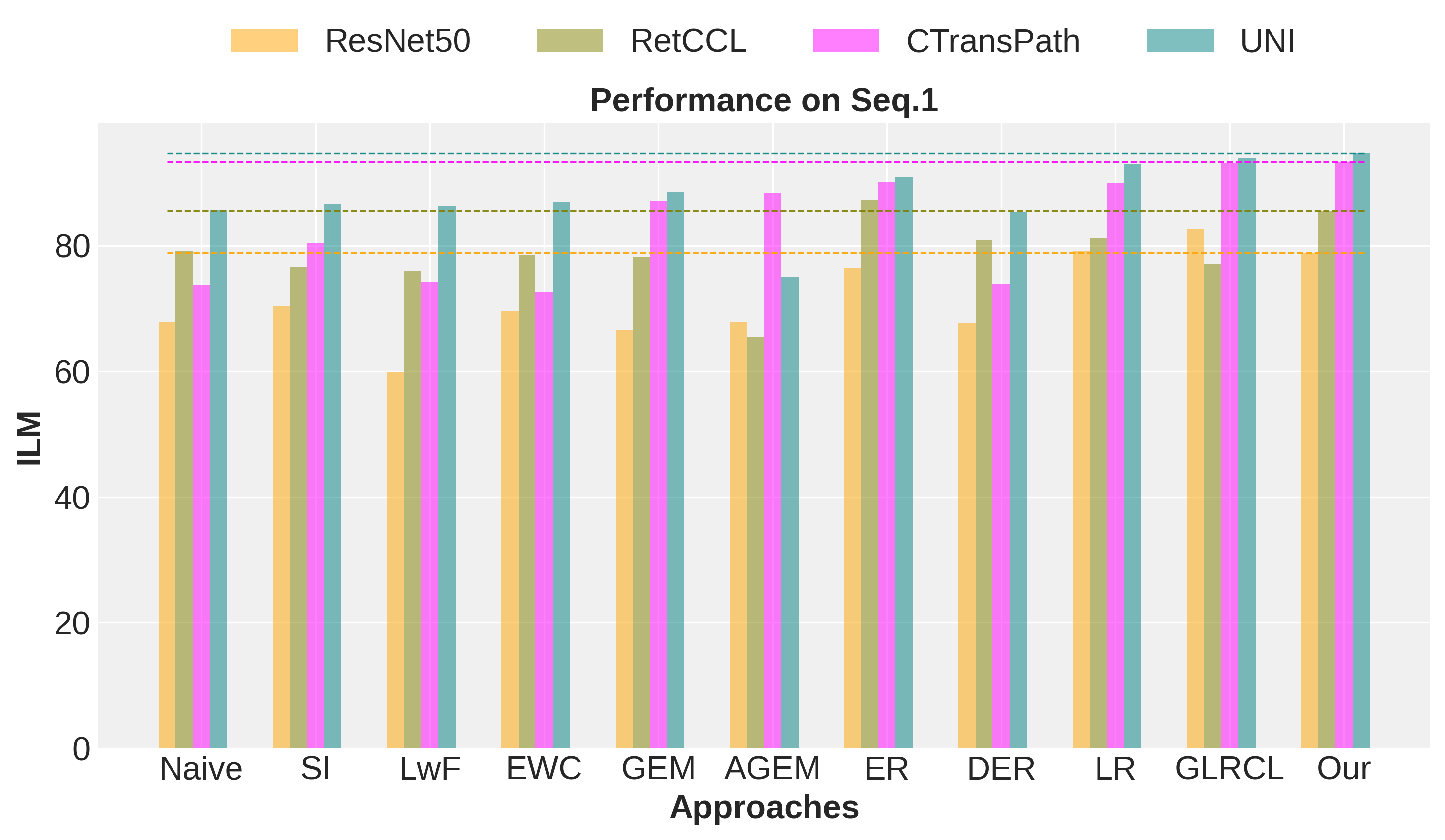}
    \includegraphics[width=0.47\textwidth]{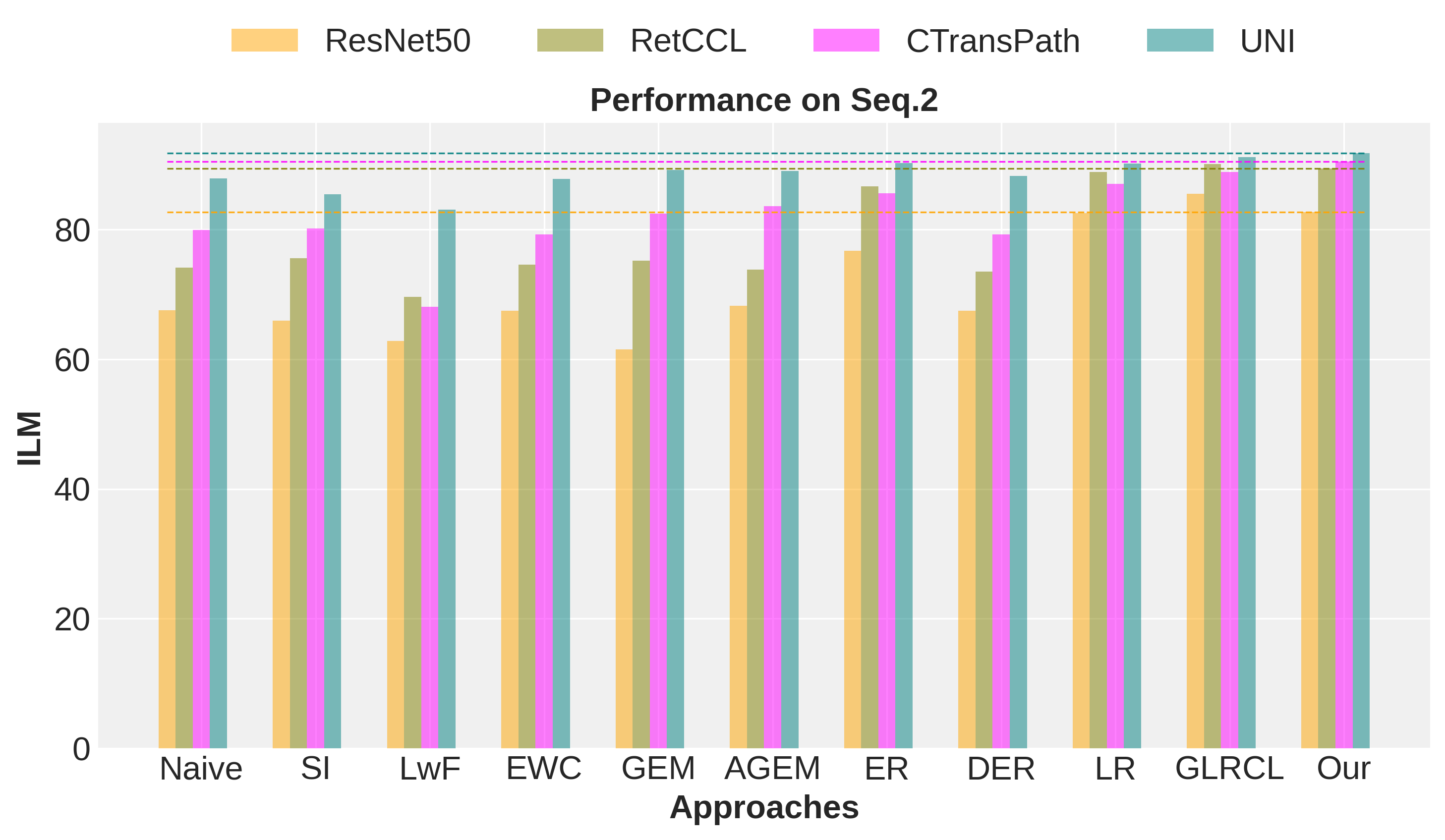}
    \includegraphics[width=0.47\textwidth]{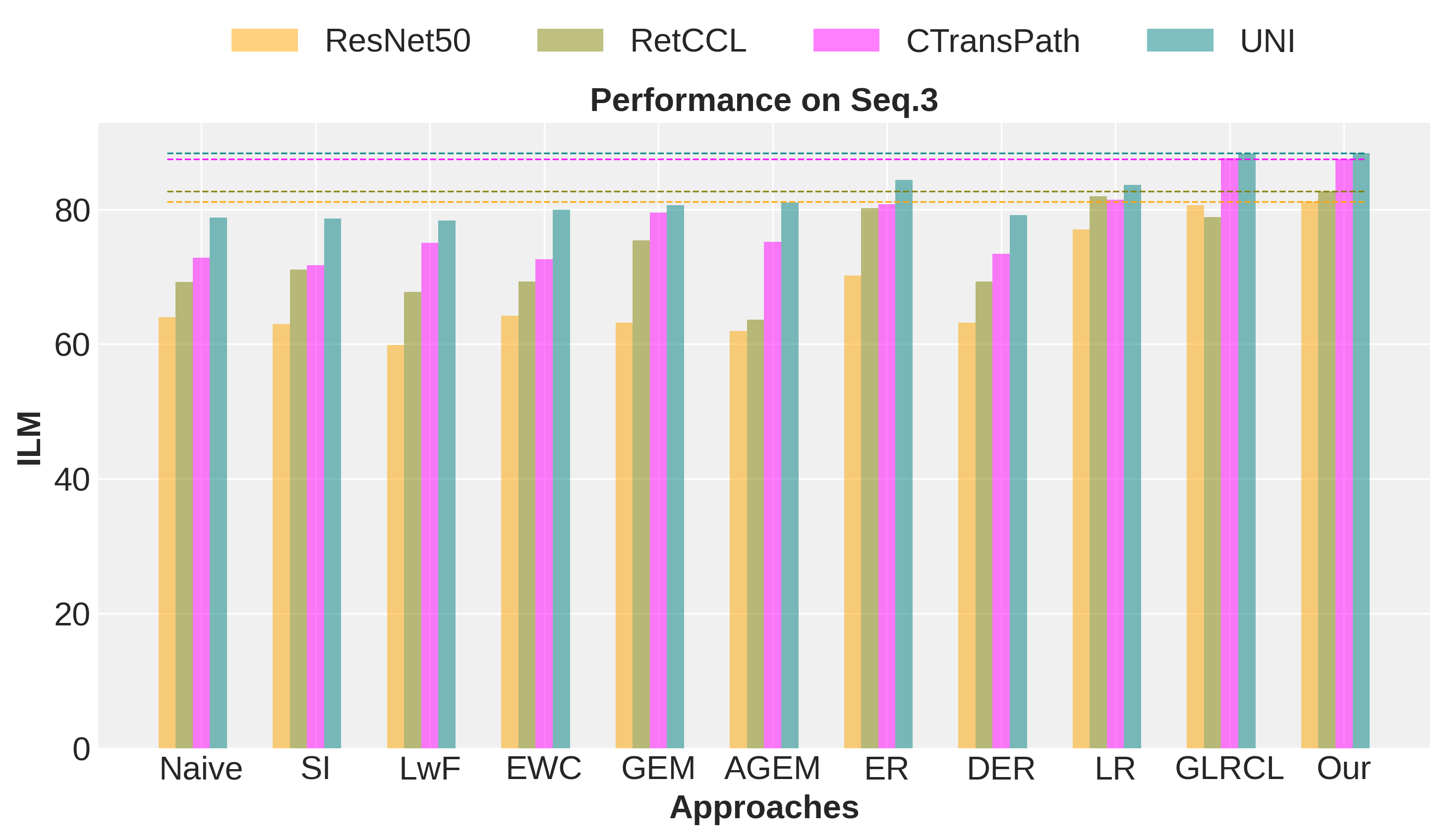}
    \includegraphics[width=0.47\textwidth]{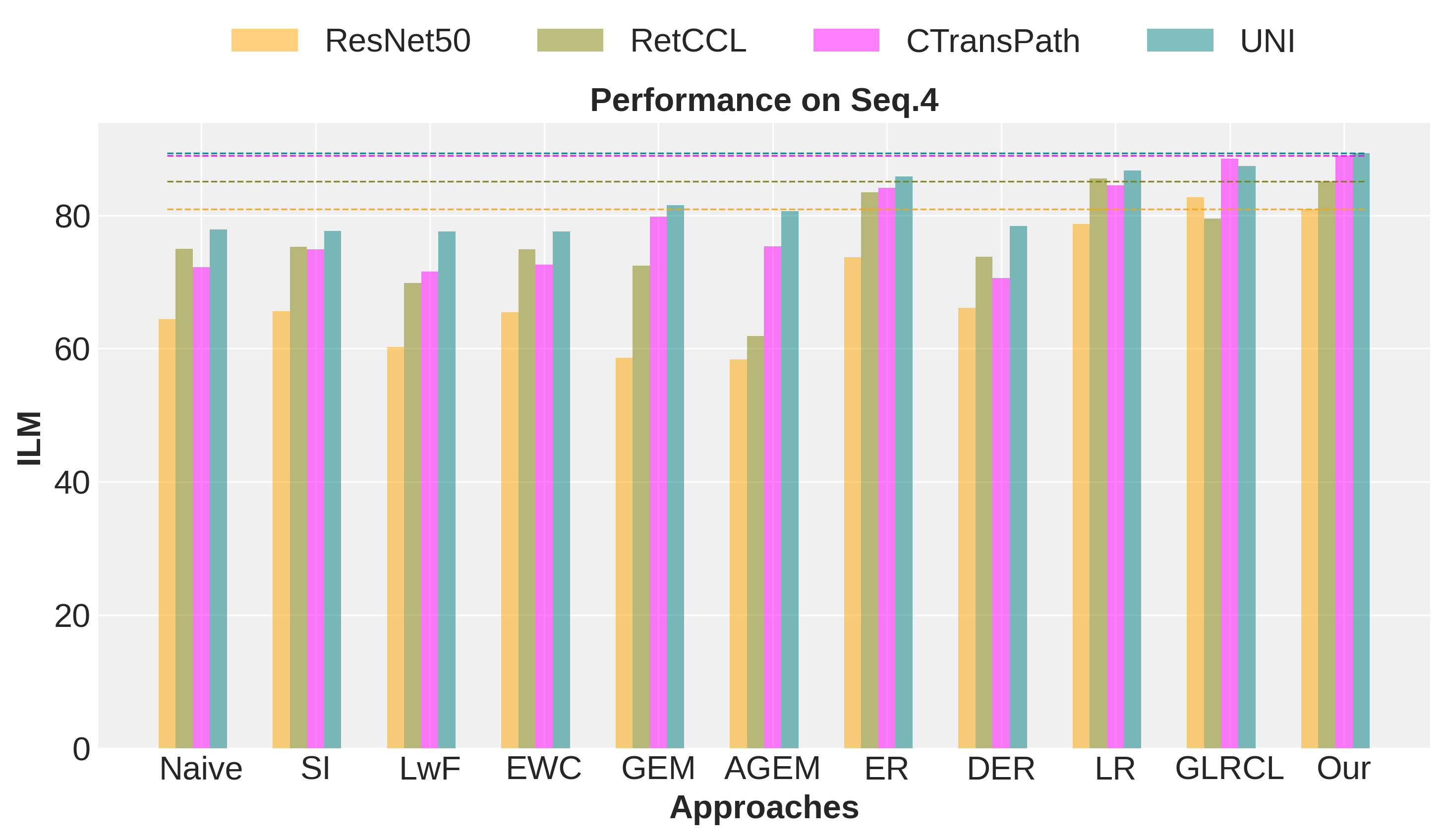}
    \caption{Backbone comparison in different approaches}
    \label{fig:barPlot}
\end{figure*}

\section{Results and Analysis}\label{sec:resultsSec}
\noindent \textbf{Catastrophic forgetting:} Here, we study the effect of catastrophic forgetting upon updating the model with new datasets in a naive fashion.
Through Fig.~{\ref{fig:naiveANDkdeResultsOrganWise}}, we analyze the performance of a dataset in different training sessions within a given sequence when using the naive and proposed learning strategy. Specifically, for a given backbone and sequence, we observe the accuracy of the first dataset after completing each training session. For a sequence with 4 datasets, this gives 4 accuracy values for the first dataset. As shown in Fig.~{\ref{fig:naiveANDkdeResultsOrganWise}}, naive learning results in significant performance degradation on the first dataset after learning new ones. In contrast, our proposed CL approach shows less fluctuation in performance, highlighting the importance of CL in mitigating catastrophic forgetting in deep models.

\noindent \textbf{Performance comparison:}
We compare our proposed approach with non-CL baselines and other state-of-the-art CL approaches based on ResNet50, RetCCL, CTransPath, and UNI backbones, as shown in \Crefrange{tab:comp_cl_all_ResNet}{tab:comp_cl_all_UNI}. These tables report ACC, ILM, and BWT metrics across all four sequences. Red represents the best performance, and blue represents the second best. Non-CL methods, such as naive learning, set the lower performance bound, while joint training achieves the highest ACC, although ILM and BWT are not applicable in that case. 
Our approach consistently achieves the highest ACC and ILM scores. However, in certain cases with weaker backbones (ResNet50 and RetCCL), it ranks second, slightly outperformed by GMM, ER, or LR. Similarly, regarding forgetting (BWT), our approach ranks first or second. It is worth noting that BWT alone can be misleading; for example in ~\Cref{tab:comp_cl_all_ResNet}, AGEM performs well in BWT, but poorly in ACC and ILM, indicating suboptimal learning. It should be noted that BWT only reflects performance changes at different points in time, not overall learning quality (diagonal values in the train-test matrix).
For the overall comparison, we also present the average ACC, ILM, and BWT values across the four sequences in the last column of~\Crefrange{tab:comp_cl_all_ResNet}{tab:comp_cl_all_UNI}. Our proposed method achieves the best ACC and ILM scores across all considered backbones, except for ResNet50, where our approach scores second best. Additionally, it significantly outperforms buffer-free approaches, indicating its superiority. 
ER or LR approaches, which store raw data in latent or image form in the buffer, may show better performance with a large buffer but with limited applicability due to storage and privacy constraints.

\noindent \textbf{Backbone Analysis:} Here, we compare different backbones for CL and naive methods using the ILM metric, summarized by bar plots in \Cref{fig:barPlot}. As expected, deep models trained on medical data outperform those trained on natural images. Foundation models with substantial exposure to medical data, such as UNI, outperform RetCCL and CTransPath. Notably, even a strong backbone (UNI) paired with a naive approach performs similarly to, or better than, a weaker backbone (ResNet50) with a CL strategy. However, despite the strength of the UNI model in standard settings, its performance declines under domain shift conditions. Our proposed approach consistently outperforms the naive method, even with the powerful UNI backbone.

\begin{figure}[!ht]
    \centering
    \includegraphics[scale=0.65]{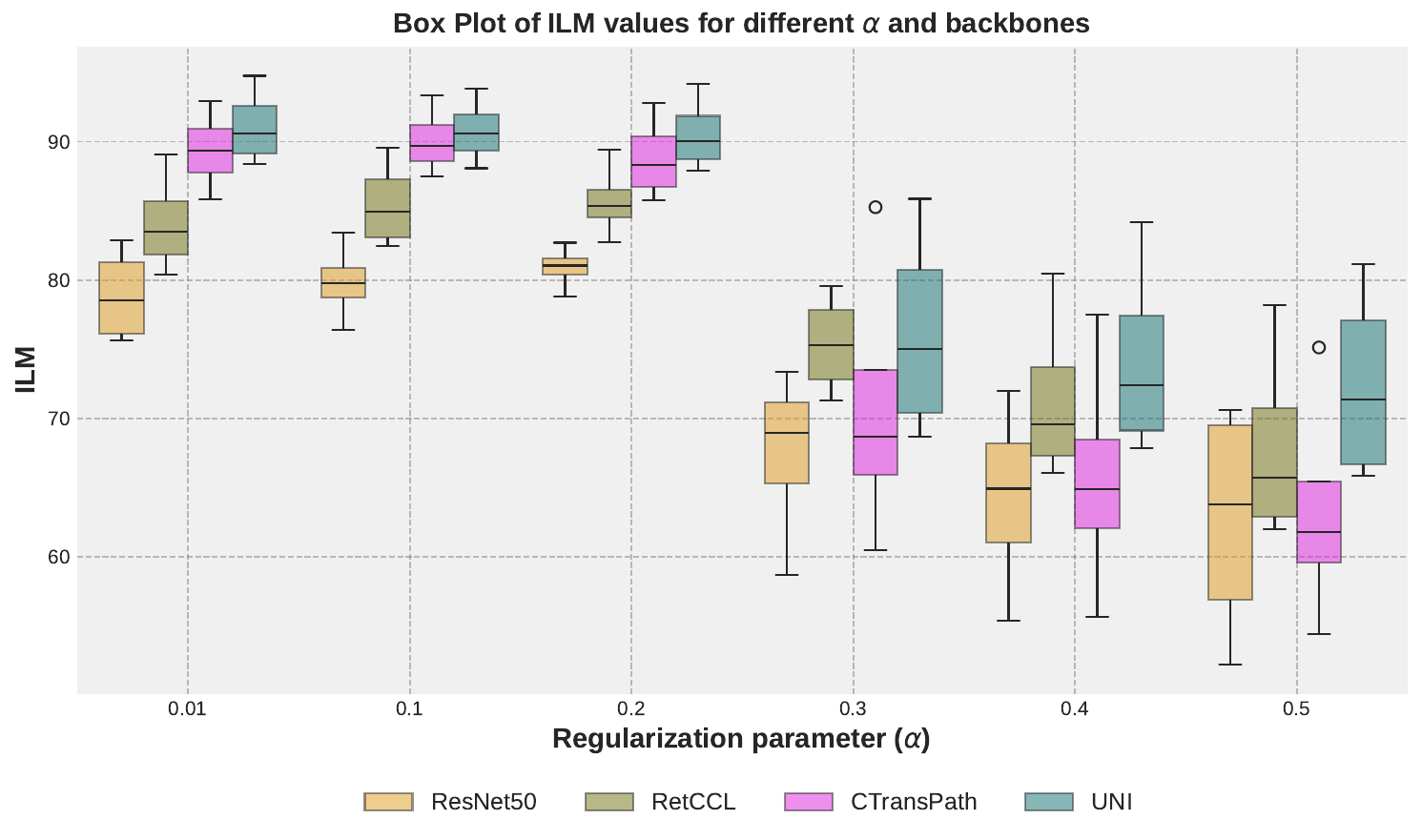}
    \caption{Performance (ILM) with different regularization factors in different backbones with the proposed approach.}
    \label{fig:alpha_study}
\end{figure}
\section{Ablation Study}

\noindent \textbf{Regularization Strength:}
Here, we analyze the impact of regularization strength (varying $\alpha$ values) on the performance of our proposed framework across multiple backbone architectures. We report the average ILM, computed for Sequences 1 to 4, in Fig.~{\ref{fig:alpha_study}}. For ResNet50 and RetCCL, the best performance is achieved at $\alpha$ = 0.2, indicating that moderate regularization effectively mitigates forgetting while maintaining plasticity for learning new datasets. In contrast, the stronger models, CTransPath and UNI, performs best at $\alpha$ = 0.1 and 0.01, respectively, suggesting that minimal regularization is sufficient to balance past knowledge retention and adaptation in conjunction with a FM. Beyond $\alpha$ = 0.2, all backbones exhibit a decline in ILM scores with increased variance, indicating that excessive regularization hinders learning new datasets. 
These findings emphasize that, while lower $\alpha$ values generally improve performance, the optimal $\alpha$ varies across architectures, underscoring the need for expert tuning to achieve the best trade-off between plasticity and stability. Overall, a small $\alpha$ is preferable for effective knowledge retention and adaptation in domain-incremental WBC classification.

\begin{table*}[!ht]
    \centering
    \caption{ Ablation study for different components of the proposed approach, i.e., distillation (DST) and KDE-based generative latent replay (GLR). Best results on average over four sequences are shown in \textbf{bold}.}
    \label{tab:abalationTable}
    \resizebox{\textwidth}{!}{%
        \def\arraystretch{0.9}
        \setlength{\tabcolsep}{5pt}
        \begin{tabular}{|c|cc|ccc|ccc|ccc|ccc|ccc|}
            \bottomrule
            & \multicolumn{2}{c|}{\cellcolor{gray!10} {\bf Approach}} & \multicolumn{3}{c|}{\cellcolor{gray!10} {\bf Seq.1}} & \multicolumn{3}{c|}{\cellcolor{gray!10} {\bf Seq.2}} & \multicolumn{3}{c|}{\cellcolor{gray!10} {\bf Seq.3}} & \multicolumn{3}{c|}{\cellcolor{gray!10} {\bf Seq.4}} & \multicolumn{3}{c|}{\cellcolor{gray!10} {\bf Average}} 
            \\
            \cline{2-18} & \cellcolor{gray!10} {\bf DST} & \cellcolor{gray!10} {\bf GLR} & \cellcolor{gray!10} {\bf ACC} & \cellcolor{gray!10} {\bf ILM} & \cellcolor{gray!10} {\bf BWT} & \cellcolor{gray!10} {\bf ACC} & \cellcolor{gray!10} {\bf ILM} & \cellcolor{gray!10} {\bf BWT} & \cellcolor{gray!10} {\bf ACC} & \cellcolor{gray!10} {\bf ILM} & \cellcolor{gray!10} {\bf BWT} & \cellcolor{gray!10} {\bf ACC} & \cellcolor{gray!10} {\bf ILM} & \cellcolor{gray!10} {\bf BWT} & \cellcolor{gray!10} {\bf ACC} & \cellcolor{gray!10} {\bf ILM} & \cellcolor{gray!10} {\bf BWT} \\
            \hline \multirow{4}{*}{\parbox[t]{3mm}{\rotatebox[origin=c]{90}{\textbf{ResNet50}}}} & \multicolumn{2}{c|}{Naive} & 56.07 {\hspace{-5pt} \tiny $\pm 1.53$} & 67.83 {\hspace{-5pt} \tiny $\pm 1.44$} & -39.13 {\hspace{-5pt} \tiny $\pm 2.54$} & 66.07 {\hspace{-5pt} \tiny $\pm 1.22$} & 67.53 {\hspace{-5pt} \tiny $\pm 0.34$} & -36.24 {\hspace{-5pt} \tiny $\pm 0.97$} & 59.77 {\hspace{-5pt} \tiny $\pm 0.68$} & 64.01 {\hspace{-5pt} \tiny $\pm 0.47$} & -29.33 {\hspace{-5pt} \tiny $\pm 1.71$} & 58.83 {\hspace{-5pt} \tiny $\pm 1.57$} & 64.45 {\hspace{-5pt} \tiny $\pm 0.82$} & -35.71 {\hspace{-5pt} \tiny $\pm 2.19$}
            & 60.18 {\hspace{-5pt} \tiny $\pm 0.36$} & 65.96 {\hspace{-5pt} \tiny $\pm 0.43$} & -35.1 {\hspace{-5pt} \tiny $\pm 0.59$} 
            \\
            & \checkmark & \ding{55} & 52.74 {\hspace{-5pt} \tiny $\pm 2.52$} & 65.26 {\hspace{-5pt} \tiny $\pm 3.30$} & -25.24 {\hspace{-5pt} \tiny $\pm 6.94$}
            & 73.73 {\hspace{-5pt} \tiny $\pm 3.67$} & 76.69 {\hspace{-5pt} \tiny $\pm 2.08$} & -15.49 {\hspace{-5pt} \tiny $\pm 4.09$}
            & 73.70 {\hspace{-5pt} \tiny $\pm 2.20$} & 73.20 {\hspace{-5pt} \tiny $\pm 1.31$} & -15.18 {\hspace{-5pt} \tiny $\pm 2.23$}
            & 71.63 {\hspace{-5pt} \tiny $\pm 0.72$} & 75.63 {\hspace{-5pt} \tiny $\pm 0.83$} & -16.40 {\hspace{-5pt} \tiny $\pm 1.44$}
            & 66.57 {\hspace{-5pt} \tiny $\pm 9.81$} & 72.02 {\hspace{-5pt} \tiny $\pm 5.27$} & -18.73 {\hspace{-5pt} \tiny $\pm 6.38$}
            \\
            & \ding{55} & \checkmark & 63.07 {\hspace{-5pt} \tiny $\pm 1.95$} & 74.67 {\hspace{-5pt} \tiny $\pm 1.33$} & -18.46 {\hspace{-5pt} \tiny $\pm 2.89$}
            & 78.60 {\hspace{-5pt} \tiny $\pm 2.66$} & 82.88 {\hspace{-5pt} \tiny $\pm 1.69$} & -9.29 {\hspace{-5pt} \tiny $\pm 2.53$}
            & 78.60 {\hspace{-5pt} \tiny $\pm 2.21$} & 77.63 {\hspace{-5pt} \tiny $\pm 1.21$} & -10.82 {\hspace{-5pt} \tiny $\pm 2.47$}
            & 82.00 {\hspace{-5pt} \tiny $\pm 1.81$} & 81.27 {\hspace{-5pt} \tiny $\pm 1.01$} & -7.09 {\hspace{-5pt} \tiny $\pm 2.06$}
            & 75.57 {\hspace{-5pt} \tiny $\pm 7.67$} & 79.11 {\hspace{-5pt} \tiny $\pm 3.46$} & -11.42 {\hspace{-5pt} \tiny $\pm 4.96$}
            \\
            & \checkmark & \checkmark & 68.73 {\hspace{-5pt} \tiny $\pm 1.54$} & 78.84 {\hspace{-5pt} \tiny $\pm 0.69$} & -11.84 {\hspace{-5pt} \tiny $\pm 0.36$}
            & 79.40 {\hspace{-5pt} \tiny $\pm 1.36$} & 82.70 {\hspace{-5pt} \tiny $\pm 0.95$} & -5.73 {\hspace{-5pt} \tiny $\pm 1.60$}
            & 83.30 {\hspace{-5pt} \tiny $\pm 1.31$} & 81.16 {\hspace{-5pt} \tiny $\pm 0.63$} & -3.80 {\hspace{-5pt} \tiny $\pm 1.47$}
            & 80.20 {\hspace{-5pt} \tiny $\pm 0.66$} & 80.93 {\hspace{-5pt} \tiny $\pm 1.18$} & -5.18 {\hspace{-5pt} \tiny $\pm 1.54$}
            & \textbf{77.91} {\hspace{-5pt} \tiny $\pm 5.64$} & \textbf{80.91} {\hspace{-5pt} \tiny $\pm 1.64$} & -6.64 {\hspace{-5pt} \tiny $\pm 3.37$}
            \\
            \cline{2-18} \multirow{4}{*}{\parbox[t]{3mm}{\rotatebox[origin=c]{90}{\textbf{RetCCL}}}} & \multicolumn{2}{c|}{Naive} & 66.23 {\hspace{-5pt} \tiny $\pm 2.21$} & 79.21 {\hspace{-5pt} \tiny $\pm 1.47$} & -26.09 {\hspace{-5pt} \tiny $\pm 2.32$} & 71.0 {\hspace{-5pt} \tiny $\pm 2.27$} & 74.15 {\hspace{-5pt} \tiny $\pm 1.05$} & -32.51 {\hspace{-5pt} \tiny $\pm 1.99$} & 58.3 {\hspace{-5pt} \tiny $\pm 1.3$} & 69.24 {\hspace{-5pt} \tiny $\pm 0.98$} & -31.15 {\hspace{-5pt} \tiny $\pm 2.28$} & 75.47 {\hspace{-5pt} \tiny $\pm 1.85$} & 74.95 {\hspace{-5pt} \tiny $\pm 0.88$} & -23.69 {\hspace{-5pt} \tiny $\pm 3.17$}
            & 67.75 {\hspace{-5pt} \tiny $\pm 0.39$} & 74.39 {\hspace{-5pt} \tiny $\pm 0.22$} & -28.36 {\hspace{-5pt} \tiny $\pm 0.44$} 
            \\
            & \checkmark & \ding{55} & 64.80 {\hspace{-5pt} \tiny $\pm 3.43$} & 76.59 {\hspace{-5pt} \tiny $\pm 3.41$} & -19.49 {\hspace{-5pt} \tiny $\pm 4.01$}
            & 85.50 {\hspace{-5pt} \tiny $\pm 1.22$} & 85.25 {\hspace{-5pt} \tiny $\pm 1.70$} & -10.22 {\hspace{-5pt} \tiny $\pm 2.71$}
            & 81.27 {\hspace{-5pt} \tiny $\pm 1.24$} & 81.49 {\hspace{-5pt} \tiny $\pm 0.32$} & -8.07 {\hspace{-5pt} \tiny $\pm 1.21$}
            & 78.80 {\hspace{-5pt} \tiny $\pm 3.66$} & 80.01 {\hspace{-5pt} \tiny $\pm 1.77$} & -14.31 {\hspace{-5pt} \tiny $\pm 1.41$}
            & 77.59 {\hspace{-5pt} \tiny $\pm 8.20$} & 80.84 {\hspace{-5pt} \tiny $\pm 3.76$} & -13.02 {\hspace{-5pt} \tiny $\pm 5.07$}
            \\
            & \ding{55} & \checkmark & 71.10 {\hspace{-5pt} \tiny $\pm 4.16$} & 79.55 {\hspace{-5pt} \tiny $\pm 4.05$} & -17.76 {\hspace{-5pt} \tiny $\pm 5.65$}
            & 86.63 {\hspace{-5pt} \tiny $\pm 1.20$} & 89.04 {\hspace{-5pt} \tiny $\pm 1.10$} & -5.60 {\hspace{-5pt} \tiny $\pm 1.36$}
            & 84.63 {\hspace{-5pt} \tiny $\pm 0.92$} & 83.23 {\hspace{-5pt} \tiny $\pm 0.95$} & -5.98 {\hspace{-5pt} \tiny $\pm 1.88$}
            & 85.30 {\hspace{-5pt} \tiny $\pm 2.15$} & 84.21 {\hspace{-5pt} \tiny $\pm 2.62$} & -9.20 {\hspace{-5pt} \tiny $\pm 4.03$}
            & 81.92 {\hspace{-5pt} \tiny $\pm 6.75$} & 84.01 {\hspace{-5pt} \tiny $\pm 4.22$} & -9.63 {\hspace{-5pt} \tiny $\pm 6.11$}
            \\
            & \checkmark & \checkmark & 78.23 {\hspace{-5pt} \tiny $\pm 2.45$} & 85.56 {\hspace{-5pt} \tiny $\pm 1.18$} & -6.62 {\hspace{-5pt} \tiny $\pm 2.05$}
            & 87.73 {\hspace{-5pt} \tiny $\pm 0.88$} & 89.43 {\hspace{-5pt} \tiny $\pm 0.99$} & -3.33 {\hspace{-5pt} \tiny $\pm 1.32$}
            & 84.00 {\hspace{-5pt} \tiny $\pm 2.21$} & 82.72 {\hspace{-5pt} \tiny $\pm 2.02$} & -5.36 {\hspace{-5pt} \tiny $\pm 2.70$}
            & 85.83 {\hspace{-5pt} \tiny $\pm 2.03$} & 85.12 {\hspace{-5pt} \tiny $\pm 1.98$} & -3.53 {\hspace{-5pt} \tiny $\pm 1.55$}
            & \textbf{83.95} {\hspace{-5pt} \tiny $\pm 4.07$} & \textbf{85.71} {\hspace{-5pt} \tiny $\pm 2.89$} & \textbf{-4.71} {\hspace{-5pt} \tiny $\pm 2.40$}
            \\
            \cline{2-18} \multirow{4}{*}{\parbox[t]{3mm}{ \rotatebox[origin=c]{90}{\textbf{\small CTransPath}}}} & \multicolumn{2}{c|}{Naive} & 57.47 {\hspace{-5pt} \tiny $\pm 3.37$} & 73.76 {\hspace{-5pt} \tiny $\pm 1.58$} & -35.51 {\hspace{-5pt} \tiny $\pm 2.67$} & 78.27 {\hspace{-5pt} \tiny $\pm 1.62$} & 79.93 {\hspace{-5pt} \tiny $\pm 1.54$} & -21.73 {\hspace{-5pt} \tiny $\pm 2.32$} & 71.93 {\hspace{-5pt} \tiny $\pm 1.66$} & 72.81 {\hspace{-5pt} \tiny $\pm 0.73$} & -26.56 {\hspace{-5pt} \tiny $\pm 1.2$} & 66.63 {\hspace{-5pt} \tiny $\pm 1.9$} & 72.21 {\hspace{-5pt} \tiny $\pm 0.82$} & -32.27 {\hspace{-5pt} \tiny $\pm 2.45$}
            & 68.58 {\hspace{-5pt} \tiny $\pm 0.72$} & 74.68 {\hspace{-5pt} \tiny $\pm 0.39$} & -29.02 {\hspace{-5pt} \tiny $\pm 0.57$} 
            \\
            & \checkmark & \ding{55} & 73.03 {\hspace{-5pt} \tiny $\pm 3.02$} & 81.77 {\hspace{-5pt} \tiny $\pm 2.16$} & -21.18 {\hspace{-5pt} \tiny $\pm 3.46$}
            & 77.17 {\hspace{-5pt} \tiny $\pm 1.15$} & 78.68 {\hspace{-5pt} \tiny $\pm 2.74$} & -21.73 {\hspace{-5pt} \tiny $\pm 3.94$}
            & 78.43 {\hspace{-5pt} \tiny $\pm 1.33$} & 76.23 {\hspace{-5pt} \tiny $\pm 1.60$} & -21.91 {\hspace{-5pt} \tiny $\pm 3.04$}
            & 73.87 {\hspace{-5pt} \tiny $\pm 3.54$} & 76.93 {\hspace{-5pt} \tiny $\pm 1.20$} & -23.87 {\hspace{-5pt} \tiny $\pm 1.66$}
            & 75.62 {\hspace{-5pt} \tiny $\pm 3.35$} & 78.40 {\hspace{-5pt} \tiny $\pm 2.94$} & -22.17 {\hspace{-5pt} \tiny $\pm 3.30$}
            \\
            & \ding{55} & \checkmark & 87.20 {\hspace{-5pt} \tiny $\pm 1.57$} & 92.44 {\hspace{-5pt} \tiny $\pm 0.76$} & -2.27 {\hspace{-5pt} \tiny $\pm 0.72$}
            & 88.57 {\hspace{-5pt} \tiny $\pm 1.54$} & 89.67 {\hspace{-5pt} \tiny $\pm 0.92$} & -3.47 {\hspace{-5pt} \tiny $\pm 1.07$}
            & 89.23 {\hspace{-5pt} \tiny $\pm 0.60$} & 87.31 {\hspace{-5pt} \tiny $\pm 1.05$} & -3.80 {\hspace{-5pt} \tiny $\pm 0.80$}
            & 88.27 {\hspace{-5pt} \tiny $\pm 1.56$} & 87.28 {\hspace{-5pt} \tiny $\pm 1.25$} & -6.13 {\hspace{-5pt} \tiny $\pm 1.84$}
            & 88.32 {\hspace{-5pt} \tiny $\pm 1.56$} & 89.17 {\hspace{-5pt} \tiny $\pm 2.35$} & -3.92 {\hspace{-5pt} \tiny $\pm 1.84$}
            \\
            & \checkmark & \checkmark & 88.20 {\hspace{-5pt} \tiny $\pm 1.81$} & 93.37 {\hspace{-5pt} \tiny $\pm 1.13$} & -1.58 {\hspace{-5pt} \tiny $\pm 1.13$}
            & 90.43 {\hspace{-5pt} \tiny $\pm 1.22$} & 90.48 {\hspace{-5pt} \tiny $\pm 1.90$} & -1.95 {\hspace{-5pt} \tiny $\pm 0.95$}
            & 89.83 {\hspace{-5pt} \tiny $\pm 1.14$} & 87.48 {\hspace{-5pt} \tiny $\pm 1.53$} & -2.62 {\hspace{-5pt} \tiny $\pm 1.34$}
            & 89.73 {\hspace{-5pt} \tiny $\pm 1.91$} & 88.95 {\hspace{-5pt} \tiny $\pm 2.52$} & -4.27 {\hspace{-5pt} \tiny $\pm 2.48$}
            & \textbf{ 89.55} {\hspace{-5pt} \tiny $\pm 1.76$} & \textbf{90.07} {\hspace{-5pt} \tiny $\pm 2.85$} & \textbf{-2.61} {\hspace{-5pt} \tiny $\pm 1.89$}
            \\
            \cline{2-18} \multirow{5}{*}{\parbox[t]{3mm}{\rotatebox[origin=c]{90}{\textbf{UNI}}}} & \multicolumn{2}{c|}{Naive} & 78.3 {\hspace{-5pt} \tiny $\pm 5.47$} & 85.69 {\hspace{-5pt} \tiny $\pm 2.39$} & -15.98 {\hspace{-5pt} \tiny $\pm 4.12$} & 86.83 {\hspace{-5pt} \tiny $\pm 1.27$} & 87.87 {\hspace{-5pt} \tiny $\pm 0.6$} & -10.18 {\hspace{-5pt} \tiny $\pm 1.58$} & 79.67 {\hspace{-5pt} \tiny $\pm 1.61$} & 78.79 {\hspace{-5pt} \tiny $\pm 0.1$} & -20.15 {\hspace{-5pt} \tiny $\pm 1.08$} & 76.43 {\hspace{-5pt} \tiny $\pm 2.48$} & 77.87 {\hspace{-5pt} \tiny $\pm 0.88$} & -22.04 {\hspace{-5pt} \tiny $\pm 2.18$}
            & 80.31 {\hspace{-5pt} \tiny $\pm 1.65$} & 82.56 {\hspace{-5pt} \tiny $\pm 0.85$} & -17.09 {\hspace{-5pt} \tiny $\pm 1.15$} 
            \\
            & \checkmark & \ding{55} & 84.80 {\hspace{-5pt} \tiny $\pm 2.02$} & 89.35 {\hspace{-5pt} \tiny $\pm 2.05$} & -10.25 {\hspace{-5pt} \tiny $\pm 3.17$}
            & 82.80 {\hspace{-5pt} \tiny $\pm 2.11$} & 86.48 {\hspace{-5pt} \tiny $\pm 1.06$} & -11.54 {\hspace{-5pt} \tiny $\pm 1.57$}
            & 80.23 {\hspace{-5pt} \tiny $\pm 0.85$} & 77.95 {\hspace{-5pt} \tiny $\pm 0.79$} & -21.04 {\hspace{-5pt} \tiny $\pm 2.08$}
            & 76.47 {\hspace{-5pt} \tiny $\pm 1.24$} & 76.83 {\hspace{-5pt} \tiny $\pm 0.97$} & -25.00 {\hspace{-5pt} \tiny $\pm 1.74$}
            & 81.08 {\hspace{-5pt} \tiny $\pm 3.52$} & 82.65 {\hspace{-5pt} \tiny $\pm 5.53$} & -16.96 {\hspace{-5pt} \tiny $\pm 6.63$}
            \\
            & \ding{55} & \checkmark
            & 89.77 {\hspace{-5pt} \tiny $\pm 1.86$} & 93.77 {\hspace{-5pt} \tiny $\pm 1.14$} & -1.98 {\hspace{-5pt} \tiny $\pm 1.16$}
            & 90.33 {\hspace{-5pt} \tiny $\pm 0.85$} & 91.41 {\hspace{-5pt} \tiny $\pm 0.99$} & -3.45 {\hspace{-5pt} \tiny $\pm 1.02$}
            & 89.23 {\hspace{-5pt} \tiny $\pm 1.26$} & 87.52 {\hspace{-5pt} \tiny $\pm 0.98$} & -4.76 {\hspace{-5pt} \tiny $\pm 2.51$}
            & 90.00 {\hspace{-5pt} \tiny $\pm 1.15$} & 89.27 {\hspace{-5pt} \tiny $\pm 0.73$} & -5.42 {\hspace{-5pt} \tiny $\pm 1.61$}
            & 89.83 {\hspace{-5pt} \tiny $\pm 1.39$} & 90.49 {\hspace{-5pt} \tiny $\pm 2.54$} & -3.90 {\hspace{-5pt} \tiny $\pm 2.13$}
            \\
            & \checkmark & \checkmark & 91.33 {\hspace{-5pt} \tiny $\pm 0.63$} & 94.75 {\hspace{-5pt} \tiny $\pm 0.38$} & -1.62 {\hspace{-5pt} \tiny $\pm 0.64$}
            & 91.07 {\hspace{-5pt} \tiny $\pm 0.85$} & 91.83 {\hspace{-5pt} \tiny $\pm 0.71$} & -2.82 {\hspace{-5pt} \tiny $\pm 0.91$}
            & 89.87 {\hspace{-5pt} \tiny $\pm 0.71$} & 88.40 {\hspace{-5pt} \tiny $\pm 0.69$} & -4.02 {\hspace{-5pt} \tiny $\pm 0.64$}
            & 89.70 {\hspace{-5pt} \tiny $\pm 1.25$} & 89.38 {\hspace{-5pt} \tiny $\pm 1.18$} & -3.27 {\hspace{-5pt} \tiny $\pm 0.86$}
            & \textbf{90.49} {\hspace{-5pt} \tiny $\pm 1.15$} & \textbf{91.09 }{\hspace{-5pt} \tiny $\pm 2.58$} & \textbf{ -2.93} {\hspace{-5pt} \tiny $\pm 1.16$}
            \\
            \toprule
        \end{tabular}
    }
\end{table*}



\noindent \textbf{Analysis of Different Components in Proposed Framework:}
Here, we compare and analyze different components of the proposed CL framework, including KDE-based generative latent replay and knowledge distillation from the past model as regularization. We present the results of our ablation study in Table~{\ref{tab:abalationTable}}, with ACC, ILM, and BWT metrics with naive and different components of our proposed approach. It can be seen that all components of our approach, both individually and in combination, outperform naive training across different sequences and backbones. The two outliers (i.e. DST-only component in Sequence 1 with ResNet50 and RetCCL) can attributed to inappropriate distillation from weak backbones. Overall, when analyzing the results showing the average over four sequences (last column in Table {\ref{tab:abalationTable}}), clearly all different components of our proposed method outperform the naive approach. Next, we analyze the contribution of each module in our approach. In the DST variant, $\text{loss}_{CE}$ and $\text{loss}_{KLD}$ losses are computed only for the current task data, and in GLR only the $\text{loss}_{CE}$ loss is used for the current task data and the generated data. The teacher model is used to obtain the label for generated data in the GLR variant. It can be observed that in all sequences and backbones, KDE-based generative latent replay (GLR) outperforms teacher-student based distillation (DST), highlighting the fact that replay-based approaches outperform regularization-based ones. However, when GLR is combined with DST (i.e. our proposed approach), it consistently outperforms individual components throughout all backbones, supporting the inclusion of DST in the proposed strategy. 
These results demonstrate the complementary benefits of knowledge distillation and generative replay, highlighting the effectiveness of our hybrid CL strategy.

\section{Conclusion}\label{sec:conc}
We introduced a KDE-based generative replay approach for sequential learning of WBC classification datasets in domain-incremental scenarios. Unlike buffer-based methods that store raw images for future replay, our approach learns KDE on latent vectors, providing a privacy-preserving solution. Through comprehensive comparison across multiple dataset sequences, our method outperforms popular CL benchmarks such as regularization, experience replay, and generative replay approaches.
Additionally, we evaluated different backbones, including ResNet50 and large-scale pre-trained models like RetCCL, CTransPath, and UNI. Even with state-of-the-art models, transfer learning on new domains causes forgetting of previously learned knowledge. Our CL strategy significantly mitigates forgetting, resulting in improved overall performance. Among all models, the UNI-based framework demonstrates the best continual learning performance.

\bibliography{0_main}

\begin{thebibliography}{10}
\urlstyle{rm}
\expandafter\ifx\csname url\endcsname\relax
  \def\url#1{\texttt{#1}}\fi
\expandafter\ifx\csname urlprefix\endcsname\relax\def\urlprefix{URL }\fi
\expandafter\ifx\csname doiprefix\endcsname\relax\def\doiprefix{DOI: }\fi
\providecommand{\bibinfo}[2]{#2}
\providecommand{\eprint}[2][]{\url{#2}}

\bibitem{goodfellow2013empirical}
\bibinfo{author}{Goodfellow, I.~J.}, \bibinfo{author}{Mirza, M.}, \bibinfo{author}{Xiao, D.}, \bibinfo{author}{Courville, A.} \& \bibinfo{author}{Bengio, Y.}
\newblock \bibinfo{journal}{\bibinfo{title}{An empirical investigation of catastrophic forgetting in gradient-based neural networks}}.
\newblock {\emph{\JournalTitle{arXiv preprint arXiv:1312.6211}}}  (\bibinfo{year}{2013}).

\bibitem{kumari2023continual}
\bibinfo{author}{Kumari, P.}, \bibinfo{author}{Chauhan, J.}, \bibinfo{author}{Bozorgpour, A.}, \bibinfo{author}{Azad, R.} \& \bibinfo{author}{Merhof, D.}
\newblock \bibinfo{journal}{\bibinfo{title}{Continual learning in medical imaging analysis: A comprehensive review of recent advancements and future prospects}}.
\newblock {\emph{\JournalTitle{arXiv preprint arXiv:2312.17004}}}  (\bibinfo{year}{2023}).

\bibitem{KUMARI2024117100}
\bibinfo{author}{Kumari, P.}, \bibinfo{author}{Choudhary, P.}, \bibinfo{author}{Kujur, V.}, \bibinfo{author}{Atrey, P.~K.} \& \bibinfo{author}{Saini, M.}
\newblock \bibinfo{journal}{\bibinfo{title}{Concept drift challenge in multimedia anomaly detection: A case study with facial datasets}}.
\newblock {\emph{\JournalTitle{Signal Processing: Image Communication}}} \bibinfo{pages}{117100} (\bibinfo{year}{2024}).

\bibitem{bhatt2022experimental}
\bibinfo{author}{Bhatt, R.}, \bibinfo{author}{Singh, S.}, \bibinfo{author}{Choudhary, P.} \& \bibinfo{author}{Saini, M.}
\newblock \bibinfo{title}{An experimental study of the concept drift challenge in farm intrusion detection using audio}.
\newblock In \emph{\bibinfo{booktitle}{2022 18th IEEE International Conference on Advanced Video and Signal Based Surveillance (AVSS)}}, \bibinfo{pages}{1--8} (\bibinfo{organization}{IEEE}, \bibinfo{year}{2022}).

\bibitem{sadafi2023continual}
\bibinfo{author}{Sadafi, A.} \emph{et~al.}
\newblock \bibinfo{journal}{\bibinfo{title}{A continual learning approach for cross-domain white blood cell classification}}.
\newblock {\emph{\JournalTitle{arXiv preprint arXiv:2308.12679}}}  (\bibinfo{year}{2023}).

\bibitem{Kum_Continual_MICCAI2024}
\bibinfo{author}{Kumari, P.} \emph{et~al.}
\newblock \bibinfo{title}{Continual domain incremental learning for privacy-aware digital pathology}.
\newblock In \emph{\bibinfo{booktitle}{International Conference on Medical Image Computing and Computer-Assisted Intervention}}, \bibinfo{pages}{34--44} (\bibinfo{organization}{Springer}, \bibinfo{year}{2024}).

\bibitem{gidel2010comparison}
\bibinfo{author}{Gidel, S.}, \bibinfo{author}{Blanc, C.}, \bibinfo{author}{Chateau, T.}, \bibinfo{author}{Checchin, P.} \& \bibinfo{author}{Trassoudaine, L.}
\newblock \bibinfo{title}{Comparison between gmm and kde data fusion methods for particle filtering: Application to pedestrian detection from laser and video measurements}.
\newblock In \emph{\bibinfo{booktitle}{2010 13th International Conference on Information Fusion}}, \bibinfo{pages}{1--7} (\bibinfo{organization}{IEEE}, \bibinfo{year}{2010}).

\bibitem{acevedo2020dataset}
\bibinfo{author}{Acevedo, A.} \emph{et~al.}
\newblock \bibinfo{journal}{\bibinfo{title}{A dataset of microscopic peripheral blood cell images for development of automatic recognition systems}}.
\newblock {\emph{\JournalTitle{Data in brief}}} \textbf{\bibinfo{volume}{30}} (\bibinfo{year}{2020}).

\bibitem{matek2019single}
\bibinfo{author}{Matek, C.}, \bibinfo{author}{Schwarz, S.}, \bibinfo{author}{Marr, C.} \& \bibinfo{author}{Spiekermann, K.}
\newblock \bibinfo{journal}{\bibinfo{title}{A single-cell morphological dataset of leukocytes from aml patients and non-malignant controls (aml-cytomorphology\_lmu)}}.
\newblock {\emph{\JournalTitle{The Cancer Imaging Archive (TCIA)[Internet]}}}  (\bibinfo{year}{2019}).

\bibitem{matek2021expert}
\bibinfo{author}{Matek, C.}, \bibinfo{author}{Krappe, S.}, \bibinfo{author}{M{\"u}nzenmayer, C.}, \bibinfo{author}{Haferlach, T.} \& \bibinfo{author}{Marr, C.}
\newblock \bibinfo{journal}{\bibinfo{title}{An expert-annotated dataset of bone marrow cytology in hematologic malignancies}}.
\newblock {\emph{\JournalTitle{The Cancer Imaging Archive}}}  (\bibinfo{year}{2021}).

\bibitem{kirkpatrick2017overcoming}
\bibinfo{author}{Kirkpatrick, J.} \emph{et~al.}
\newblock \bibinfo{journal}{\bibinfo{title}{Overcoming catastrophic forgetting in neural networks}}.
\newblock {\emph{\JournalTitle{Proceedings of the national academy of sciences}}} \textbf{\bibinfo{volume}{114}}, \bibinfo{pages}{3521--3526} (\bibinfo{year}{2017}).

\bibitem{zenke2017continual}
\bibinfo{author}{Zenke, F.}, \bibinfo{author}{Poole, B.} \& \bibinfo{author}{Ganguli, S.}
\newblock \bibinfo{title}{Continual learning through synaptic intelligence}.
\newblock In \emph{\bibinfo{booktitle}{International conference on machine learning}}, \bibinfo{pages}{3987--3995} (\bibinfo{organization}{PMLR}, \bibinfo{year}{2017}).

\bibitem{radio3}
\bibinfo{author}{Li, Z.} \& \bibinfo{author}{Hoiem, D.}
\newblock \bibinfo{journal}{\bibinfo{title}{Learning without forgetting}}.
\newblock {\emph{\JournalTitle{IEEE Transactions on Pattern Analysis and Machine Intelligence}}} \textbf{\bibinfo{volume}{40}}, \bibinfo{pages}{2935--2947}, \doiprefix\url{10.1109/TPAMI.2017.2773081} (\bibinfo{year}{2018}).

\bibitem{buzzega2020dark}
\bibinfo{author}{Buzzega, P.}, \bibinfo{author}{Boschini, M.}, \bibinfo{author}{Porrello, A.}, \bibinfo{author}{Abati, D.} \& \bibinfo{author}{Calderara, S.}
\newblock \bibinfo{journal}{\bibinfo{title}{Dark experience for general continual learning: a strong, simple baseline}}.
\newblock {\emph{\JournalTitle{Advances in neural information processing systems}}} \textbf{\bibinfo{volume}{33}}, \bibinfo{pages}{15920--15930} (\bibinfo{year}{2020}).

\bibitem{rolnick2019experience}
\bibinfo{author}{Rolnick, D.}, \bibinfo{author}{Ahuja, A.}, \bibinfo{author}{Schwarz, J.}, \bibinfo{author}{Lillicrap, T.} \& \bibinfo{author}{Wayne, G.}
\newblock \bibinfo{journal}{\bibinfo{title}{Experience replay for continual learning}}.
\newblock {\emph{\JournalTitle{Advances in Neural Information Processing Systems}}} \textbf{\bibinfo{volume}{32}} (\bibinfo{year}{2019}).

\bibitem{pellegrini2020latent}
\bibinfo{author}{Pellegrini, L.}, \bibinfo{author}{Graffieti, G.}, \bibinfo{author}{Lomonaco, V.} \& \bibinfo{author}{Maltoni, D.}
\newblock \bibinfo{title}{Latent replay for real-time continual learning}.
\newblock In \emph{\bibinfo{booktitle}{2020 IEEE/RSJ International Conference on Intelligent Robots and Systems (IROS)}}, \bibinfo{pages}{10203--10209} (\bibinfo{organization}{IEEE}, \bibinfo{year}{2020}).

\bibitem{lopez2017gradient}
\bibinfo{author}{Lopez-Paz, D.} \& \bibinfo{author}{Ranzato, M.}
\newblock \bibinfo{journal}{\bibinfo{title}{Gradient episodic memory for continual learning}}.
\newblock {\emph{\JournalTitle{Advances in neural information processing systems}}} \textbf{\bibinfo{volume}{30}} (\bibinfo{year}{2017}).

\bibitem{chaudhry2018efficient}
\bibinfo{author}{Chaudhry, A.}, \bibinfo{author}{Ranzato, M.}, \bibinfo{author}{Rohrbach, M.} \& \bibinfo{author}{Elhoseiny, M.}
\newblock \bibinfo{journal}{\bibinfo{title}{Efficient lifelong learning with a-gem}}.
\newblock {\emph{\JournalTitle{arXiv preprint arXiv:1812.00420}}}  (\bibinfo{year}{2018}).

\bibitem{wang2023retccl}
\bibinfo{author}{Wang, X.} \emph{et~al.}
\newblock \bibinfo{journal}{\bibinfo{title}{Retccl: Clustering-guided contrastive learning for whole-slide image retrieval}}.
\newblock {\emph{\JournalTitle{Medical image analysis}}} \textbf{\bibinfo{volume}{83}}, \bibinfo{pages}{102645} (\bibinfo{year}{2023}).

\bibitem{wang2022transformer}
\bibinfo{author}{Wang, X.} \emph{et~al.}
\newblock \bibinfo{journal}{\bibinfo{title}{Transformer-based unsupervised contrastive learning for histopathological image classification}}.
\newblock {\emph{\JournalTitle{Medical image analysis}}} \textbf{\bibinfo{volume}{81}}, \bibinfo{pages}{102559} (\bibinfo{year}{2022}).

\bibitem{chen2024towards}
\bibinfo{author}{Chen, R.~J.} \emph{et~al.}
\newblock \bibinfo{journal}{\bibinfo{title}{Towards a general-purpose foundation model for computational pathology}}.
\newblock {\emph{\JournalTitle{Nature Medicine}}} \textbf{\bibinfo{volume}{30}}, \bibinfo{pages}{850--862} (\bibinfo{year}{2024}).

\bibitem{koch2024dinobloom}
\bibinfo{author}{Koch, V.} \emph{et~al.}
\newblock \bibinfo{title}{Dinobloom: A foundation model for generalizable cell embeddings in hematology}.
\newblock In \emph{\bibinfo{booktitle}{International Conference on Medical Image Computing and Computer-Assisted Intervention}}, \bibinfo{pages}{520--530} (\bibinfo{organization}{Springer}, \bibinfo{year}{2024}).

\bibitem{avalanche}
\bibinfo{author}{Lomonaco, V.} \emph{et~al.}
\newblock \bibinfo{title}{Avalanche: an end-to-end library for continual learning}.
\newblock In \emph{\bibinfo{booktitle}{Proceedings of IEEE Conference on Computer Vision and Pattern Recognition}}, 2nd Continual Learning in Computer Vision Workshop (\bibinfo{year}{2021}).

\bibitem{diaz2018don}
\bibinfo{author}{D{\'\i}az-Rodr{\'\i}guez, N.}, \bibinfo{author}{Lomonaco, V.}, \bibinfo{author}{Filliat, D.} \& \bibinfo{author}{Maltoni, D.}
\newblock \bibinfo{journal}{\bibinfo{title}{Don't forget, there is more than forgetting: new metrics for continual learning}}.
\newblock {\emph{\JournalTitle{arXiv preprint arXiv:1810.13166}}}  (\bibinfo{year}{2018}).

\end{thebibliography}

\section*{Acknowledgements}
This work was supported by the German Research Foundation (Deutsche Forschungsgemeinschaft, DFG) under project number 527820737. The authors gratefully acknowledge the computational and data resources provided by the Leibniz Supercomputing Centre (www.lrz.de).

\section*{Authors contributions statement}
P.K. and A.B. conceived and conducted the experiments. P.K., A.B., D.R., and D.M. analyzed the results. E.J., M.C. and C.M. provided clinical feedback. All authors reviewed the manuscript.

\section*{Additional information}
\textbf{Competing interests} The authors declare that they have no conflicts of interest related to this work.

\section*{Data availability}
The PBC, LMU, and MLL datasets used during the current study are publicly available for research purposes.
The UKA dataset analyzed during the current study is currently not publicly available, since there is no ethics approval available that would allow to publicly share this medical dataset. Edgar Jost (ejost@ukaachen.de) should be contacted if someone would like to request the "UKA dataset" data from this study.

\end{document}